\RequirePackage{fix-cm}
\documentclass[twocolumn]{svjour3}          
\smartqed  
\usepackage{graphicx}

\usepackage{moreverb,url}

\usepackage[colorlinks,bookmarksopen,bookmarksnumbered,citecolor=red,urlcolor=red]{hyperref}

\usepackage{graphicx}
\usepackage{gensymb}
\usepackage[table]{xcolor}
\usepackage{multirow}
\usepackage{enumerate}
\usepackage{array,booktabs,arydshln}
\usepackage{graphicx}
\usepackage[acronym,nomain]{glossaries}

\usepackage{amsmath}
\usepackage{amssymb}

\DeclareMathOperator*{\argmin}{argmin}

\makeglossaries
\begin{document}

\title{Motion Planning and Control for Mobile Robot Navigation Using Machine Learning: a Survey
}


\author{Xuesu Xiao, Bo Liu, Garrett Warnell, and Peter Stone}


\institute{Xuesu Xiao, Bo Liu, and Peter Stone \at
              Department of Computer Science,
        University of Texas at Austin, Austin, Texas 78712. \\Peter Stone is also with Sony AI. 
              \email{\{xiao, bliu, pstone\}@cs.utexas.edu}           
           \and
           Garrett Warnell \at
           Computational and Information Sciences Directorate, Army Research Laboratory, Austin, Texas 78712.
           \email{garrett.a.warnell.civ@mail.mil}
}

\date{Received: date / Accepted: date}

\maketitle


\begin{abstract}
Moving in complex environments is an essential capability of intelligent mobile robots. Decades of research and engineering have been dedicated to developing sophisticated navigation systems to move mobile robots from one point to another. Despite their overall success, a recently emerging research thrust is devoted to developing machine learning techniques to address the same problem, based in large part on the success of deep learning. However, to date, there has not been much direct comparison between the classical and emerging paradigms to this problem. In this article, we survey recent works that apply machine learning for motion planning and control in mobile robot navigation, within the context of classical navigation systems. The surveyed works are classified into different categories, which delineate the relationship of the learning approaches to classical methods. Based on this classification, we identify common challenges and promising future directions.  
\end{abstract}

\section{INTRODUCTION}
\label{sec::introduction}

Autonomous mobile robot navigation---i.e., the ability of an artificial agent to move itself towards a specified waypoint smoothly and without collision---has attracted a large amount of attention and research, leading to hundreds of approaches over several decades. Autonomous navigation lies at the intersection of many different fundamental research areas in robotics. Moving the robot from one position to another requires techniques in perception, state estimation (e.g., localization, mapping, and world representation), path planning, and motion planning and control. While many sophisticated autonomous navigation systems have been proposed, they usually follow a classical hierarchical planning paradigm: global path planning combined with local motion control. While this paradigm has enabled autonomous navigation on a variety of different mobile robot platforms---including unmanned ground, aerial, surface, and underwater vehicles---its typical performance still lags that which can be achieved through human teleoperation. For example, mobile robots still cannot achieve robust navigation in highly constrained or highly diverse environments. 

With the recent explosion in machine learning research, data-driven techniques---in stark contrast to the classical hierarchical planning framework---are also being applied to the autonomous navigation problem. Early work of this variety has produced systems that use end-to-end learning algorithms in order to find navigation systems that map directly from perceptual inputs to motion commands, thus bypassing the classical hierarchical paradigm. These systems allow a robot to navigate without any other traditional symbolic, rule-based human knowledge or engineering design. However, these methods have thus far proven to be extremely data-hungry and are typically unable to provide any safety guarantees. Moreover, they lack explainability when compared to the classical approaches. For example, after being trained on hundreds of thousands of instances of training data, during deployment, the cause of a collision may still not be easily identifiable. This lack of explainability makes further debugging and improvement from this collision difficult. As a result, many learning-based approaches have been studied mostly in simulation and have only occasionally been applied to simple real-world environments as ``proof-of-concept'' systems for learned navigation.

These two schools of thought---classical hierarchical planning and machine learning---each seek to tackle the same problem of autonomous navigation, but as of yet, have not been compared systematically. Such a comparison would be useful given that each class of approach typically dominates in its own, largely separate community. The robotics community has gravitated toward classical hierarchical planners due to their ability to reliably deploy such systems in the real world, but many are perhaps unaware of recent advances in machine learning for navigation. Similarly, while the machine learning community is indeed developing new, potentially powerful data-driven approaches, they are often only compared against previous learning approaches, thus neglecting the most pressing challenges for real-world navigation. A lack of awareness of, and thus a failure to address, these challenges makes it less likely for roboticists to consider their methods for deployment in real-world environments.

In this article, we provide exactly the comparison described above based on existing literature on machine learning for motion planning and control\footnote{In mobile robot navigation, ``motion planning" mostly focuses on relatively long-term sequences of robot positions, orientations, and their high-order derivatives, while motion control generally refers to relatively low-level motor commands, e.g., linear and angular velocities. However, the line between them is blurry, and we do not adhere to any strict distinction in terminology in this survey.} in mobile robot navigation. We qualitatively compare and contrast recent machine learning navigation approaches with the classical approaches from the navigation literature.
Specifically, we adopt a robotics perspective and review works that use machine learning for motion planning and control in mobile robot navigation.
By examining machine learning work in the context of the classical navigation problem, we reveal the relationship between those learning methods and existing classical methods along several different dimensions, including functional scope (e.g., whether learning is end-to-end, or focused on a specific individual sub task) and navigation performance. 

The goals of this survey are to highlight recent advances in applying learning methods to robot navigation problems, to organize them by their role in the overall navigation pipeline, to discuss their strengths and weaknesses, and to identify interesting future directions for research.
The article is organized as follows:

\begin{itemize}
    \item Section \ref{sec::navigation} provides a brief background of classical mobile robot navigation;
    \item Section \ref{sec::learning} contains the main content, which is further divided into three subsections discussing learning methods that (i) completely replace the entire classical navigation pipeline, (ii) replace only a navigation subsystem, and (iii) adopt learning to improve upon existing components of a navigation stack;
    \item Section \ref{sec::comparison} re-categorizes the same papers in Section \ref{sec::learning}, comparing the performance between the discussed methods and the classical approaches with respect to five performance categories;
    \item Section \ref{sec::taxanomies} discusses the application domains and the input modalities of these learning methods, based on the same pool of papers;
    \item Section \ref{sec::insights} provides analyses, challenges, and future research directions that the authors think will be of value. 
\end{itemize}

Our survey differs from previous surveys on machine learning for robot motion \cite{tai2016deep1} \cite{tai2016survey}. In particular, Tai and Liu \cite{tai2016deep1} surveyed deep learning methods for mobile robots from perception to control systems. Another relevant survey \cite{tai2016survey} provided a comprehensive overview on what kinds of deep learning methods have been used for robot control, from reinforcement learning to imitation learning, and how they have varied along the learning dimension. In contrast to these two surveys from the learning perspective, our work focuses on categorizing existing learning works based on their relationships with respect to the classical navigation pipeline. We focus on works that we believe to have significant relevance to real robot navigation. In other words, we only include machine learning approaches for motion planning and control that actually \emph{move} mobile robots in their (simulated or physical) environments, excluding papers that, for example, use machine learning to navigate (non-robotic) agents in gridworld-like environments \cite{NEURIPS2020_985e9a46}, address semantic mapping but without using the learned semantics to move mobile robots \cite{jiang2021rellis}, or focus on robotic manipulation tasks \cite{kroemer2021review}. In short, we aim to provide the robotics community with a summary of which classical navigation problems are worth examining from a machine learning perspective, which have already been investigated, and which have yet to be explored. For the learning community, we aim to highlight the mobile robot navigation problems that remain unsolved by the classical approaches despite decades of research and engineering effort. 

\vspace{12pt}

\section{CLASSICAL MOBILE ROBOT NAVIGATION}
\label{sec::navigation}

\newacronym{sfm}{SfM}{Structure from Motion}
\newacronym{slam}{SLAM}{Simultaneous Localization And Mapping}
\newacronym{vo}{VO}{Visual Odometry}
\newacronym{prm}{PRM}{Probabilistic Road Maps}
\newacronym{rrt}{RRT}{Rapidly-exploring Random Trees}
\newacronym{ros}{ROS}{Robot Operating System}

In general, the classical mobile robot navigation problem is to generate a sequence of motion commands $A^* = \{a_0, a_1, a_2, ... \}$ to move a robot from its current start location $s$ to a desired goal location $g$ in a given environment $E$: 
\begin{equation}
A^* = \argmin_{A \in \mathcal{A}} J(E, s, g), 
\end{equation}
where $\mathcal{A}$ is the space of all possible sequences of motion commands and $J(\cdot)$ is a cost function that the navigation system aims to minimize. 
The actions can be linear and angular velocities for differential-drive robots, steering angle for Ackermann-steer vehicles, propeller thrust for quadrotors, etc. 
The environment $E$, either in 2D or 3D for ground or aerial navigation, is instantiated by leveraging the robot's sensory perceptions (e.g., LiDAR, camera, wheel encoder, inertial sensor) and/or pre-built maps of the environment. Depending on the navigation scenario, $J$ can take a variety of forms, which can include the robot's motion constraints, obstacle avoidance, shortest path or time, social compliance, offroad stability, and/or other domain-dependent measures. Since many navigation problems cover a large physical space and it is computationally infeasible to generate fine-grained motion sequences over long horizons, most classical navigation systems decompose $J$ and tackle it in a hierarchical manner. We use the following analogy to human navigation to illustrate such hierarchy.

Consider the problem faced by a person who would like to navigate from her bedroom to the neighborhood park. She needs to first come up with a coarse plan connecting her current location (i.e., the bedroom) to the park. That may include going to the living room through the bedroom door, following the hallway leading to the dining room, exiting the house and turning right, walking straight until the second intersection, and then turning right into the park. This sequence of high-level steps is based upon a good representation of the world, such as knowing the layout of the house and the map of the neighborhood. When the person starts walking, she may need to go around some toys left in the hallway, greet her partner in the dining room, avoid a new construction site on the sidewalk, and/or yield to a car at a red light. These behaviors are not planned beforehand, but rather are reactions to immediate situations.

The classical robotics approach to navigation follows a similar decomposition of the overall navigation task.  In particular, the bottom of Figure \ref{tab::nav_table} gives an overview of the classical mobile navigation pipeline. Based on a specific goal, the perception is processed to form a global representation of the world, which could either be based on recent perceptions or a given map. Passing the goal and perception into a parameterized global planner, a global path is produced, usually in the form of a sequence of local goals. This process corresponds to the human's coarse plan. Taking the closest local goal, the robot then uses the output of perception to create a local representation. A parameterized local planner then is responsible for generating motion commands, defined as inputs directly fed to motor controllers to generate raw motor commands, which are both safe and goal-oriented. Note that the perception may come not only from ego-centric onboard sensors, but also from other exterior systems such as GPS, acoustic transponders, or motion capture systems. This section introduces both the global and local reasoning levels of the classical mobile robot navigation pipeline. 

\vspace{10pt}

\subsection{Global Planning}
Global planning aims to find a coarse path leading from the current position to the final goal. The workspace of mobile robots is defined as the geometric space where the robot moves, and usually resides in special Euclidean group, \emph{SE(2)} or \emph{SE(3)} \cite{lavalle2006planning}. The global goal is specified as the final \emph{pose}, i.e., a combination of position and orientation. 

The global representation is reconstructed using both current and previous perception streams. Prior knowledge can also be utilized, such as an existing map. Common perception modalities include sonar, ultrasonic sensors, LiDAR, and, more popular recently, vision. The raw perception streams are processed using classical signal processing techniques, such as filtering, noise suppression, and feature extraction, and then used to reconstruct a global representation of the environment. Techniques such as \gls{sfm} \cite{ullman1979interpretation}, \gls{slam} \cite{durrant2006simultaneous}, and \gls{vo} \cite{nister2004visual} are used to generate the world representation from sensor data. 

Roboticists usually construct global representations with relatively sparse structures, such as graphs. \gls{prm} \cite{kavraki1996probabilistic} and \gls{rrt} \cite{lavalle1998rapidly} are mature sampling-based techniques for tackling large-scale and fine-resolution (usually continuous) workspaces in a timely manner. If the workspace is relatively small or low-resolution, a coarse occupancy grid \cite{elfes1989using} or costmap \cite{jaillet2010sampling} \cite{lu2014layered} may suffice. Occupancy grids treat the workspace as a grid of occupied or unoccupied cells, while costmaps construct continuous cost values for these cells. Note that the global representation may omit details such as obstacles or fine structures, since these can be handled by local planning and may also be subject to change. Existing knowledge or belief about the world, especially regions beyond the current perceptual range, can be used in the representation as well, and they are updated or corrected when the robot reaches these unknown regions. 

Based on a coarse global representation and an evaluation criterion, such as minimum distance, global planners find a reasonable path connecting the current configuration to the global goal. To be computationally efficient, global planners usually assume the robot is a point mass and only consider its position, not its orientation. Search-based algorithms are appropriate for this task. For example, a shortest-path algorithm could be used on a \gls{prm} representation, occupancy grid, or costmap. Example algorithms include depth- or breadth-first search, Dijkstra's algorithm \cite{dijkstra1959note}, and A* search with heuristics \cite{Hart1968}. Note that if the global representation changes due to new observations, the global planner needs to replan quickly. Algorithms such as D* Lite \cite{koenig2002d} can be suitable for such dynamic replanning.

As the output of global reasoning, the global planner generates a coarse global path, usually in terms of a sequence of waypoints, and passes it along to the local planning phase. The local planner is then in charge of generating motions to execute this path. 

\vspace{10pt}

\subsection{Local Planning}
Generating low-level motion commands requires a more sophisticated model of the mobile robot (either kinematic or dynamic) and a more accurate local representation of the world, as compared to the global level. When deployed online, these models often require significant computation. Therefore local planners typically only reason about the robot's immediate vicinity. 

Local planners take a local goal as input, which is typically provided by the global planner in the form of either a single position relatively close to the robot in the workspace, or a short segment of the global path starting from the robot. It is the local planner's responsibility to generate kinodynamically-feasible motion commands either to move the robot to the single local goal within an acceptable radius or to follow the segment of the global path. The local planner then moves on to the next local goal or the global path segment gets updated. 

The local environment representation is usually constructed by the current perceptual stream only, unless computation and memory resources allow for incorporating some history information. This representation needs to be more precise than the global one since it directly influences the robot's actual motion and can therefore impact safety. Common representations are occupancy grids \cite{elfes1989using} and costmaps \cite{jaillet2010sampling} \cite{lu2014layered}, but with a finer resolution and smaller footprint when compared to the representation used by the global planner. 

Local planners typically require both the surrounding world representation and a model of the robot, e.g., holonomic, differential drive, or Ackerman-steering vehicle models. Aiming at optimizing a certain objective, such as distance to local goal and clearance from obstacles, local planners compute motion commands to move the robot toward the local goal. At the same time, they need to respect the surroundings, e.g., following the road or, most importantly, avoiding obstacles \cite{quinlan1993elastic} \cite{fox1997dynamic}. The output of the local planner is either discrete high-level commands, such as \texttt{turn 45\degree~left}, or continuous linear and angular velocities. These motion commands are finally fed into low-level motor controllers. 

\begin{figure}
\centering
\includegraphics[width=1\columnwidth]{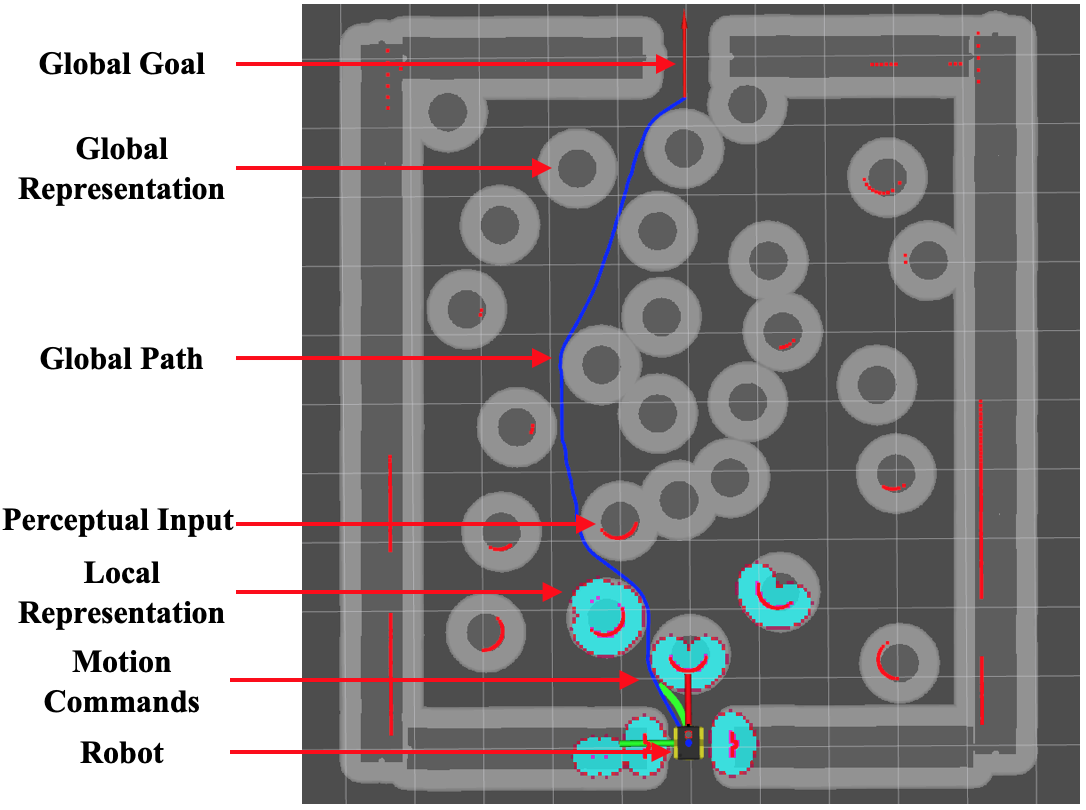}
\caption{Classical Navigation System: Given a global goal (red arrow indicating the final pose) and a global representation (grey area, based on a pre-built map or the robot's memory), the global planner produces a global path (blue line). Using the perceptual input (red dots indicating LiDAR returns), a finer and smaller local representation (cyan area) is built. Based on all this information, current motion commands (green line) are computed to move the robot \cite{ros_move_base}. }
\label{fig::lp_gp}
\end{figure}

An illustration of an example classical navigation system, \gls{ros} \texttt{move\_base} \cite{ros_move_base}, is shown in Figure \ref{fig::lp_gp}.

\vspace{12pt}
\section{SCOPE OF LEARNING FOR NAVIGATION}
\label{sec::learning}

\newacronym{il}{IL}{Imitation Learning}
\newacronym{gru}{GRU}{Gated Recurrent Units}
\newacronym{macn}{MACN}{Memory Augmented Control Network}
\newacronym{drl}{DRL}{Deep Reinforcement Learning}
\newacronym{dqn}{DQN}{Deep Q-Network}
\newacronym{cnn}{CNN}{Convolutional Neural Network}
\newacronym{a3c}{A3C}{Asynchronous Advantage Actor-Critic}
\newacronym{rl}{RL}{Reinforcement Learning}
\newacronym{vin}{VIN}{Value Iteration Network}
\newacronym{bc}{BC}{Behavior Cloning}
\newacronym{sacadrl}{SA-CADRL}{Socially Aware CADRL}
\newacronym{ddpg}{DDPG}{Deep Deterministic Policy Gradient}
\newacronym{orca}{ORCA}{Optimal Reciprocal Collision Avoidance}
\newacronym{alvinn}{ALVINN}{Autonomous Land Vehicle In a Neural Network}
\newacronym{lagr}{LAGR}{Learning Applied to Ground Robots}
\newacronym{mpc}{MPC}{Model Predictive Controller}
\newacronym{dagger}{DAgger}{Dataset Aggregation}
\newacronym{uav}{UAV}{Unmanned Aerial Vehicle}
\newacronym{lfh}{LfH}{Learning from Hallucination}
\newacronym{dwa}{DWA}{Dynamic Window Approach}
\newacronym{gem}{GEM}{Gradient Episodic Memory}
\newacronym{mbrl}{MBRL}{Model-Based Reinforcement Learning}
\newacronym{mocap}{MoCap}{Motion Capture}
\newacronym{irl}{IRL}{Inverse Reinforcement Learning}
\newacronym{gpmp2}{GPMP2}{Gaussian Process Motion Planning 2}
\newacronym{anfis}{ANFIS}{Artificial Neuro-Fuzzy Inference System}
\newacronym{appl}{APPL}{Adaptive Planner Parameter Learning}
\newacronym{e-band}{E-Band}{Elastic-Bands}
\newacronym{scn}{SCN}{Socially Concomitant Navigation}
\newacronym{hmm}{HMM}{Hidden Markov Model}

Despite the success of using conventional approaches to tackle motion planning and control for mobile robot navigation, these approaches still require an extensive amount of engineering effort before they can reliably be deployed in the real world. As a way to potentially reduce this human engineering effort, many navigation methods based on machine learning have been recently proposed in the literature. Section \ref{sec::learning} surveys relevant papers in which the scope of work may be categorized as using machine learning to solve problems that arise in the context of the classical mobile robot navigation pipeline described in Section \ref{sec::navigation}.

\begin{figure*}
\centering
\caption{Classical Navigation Pipeline}
\label{tab::nav_table}
\rotatebox{90}{\includegraphics[width=1\textheight]{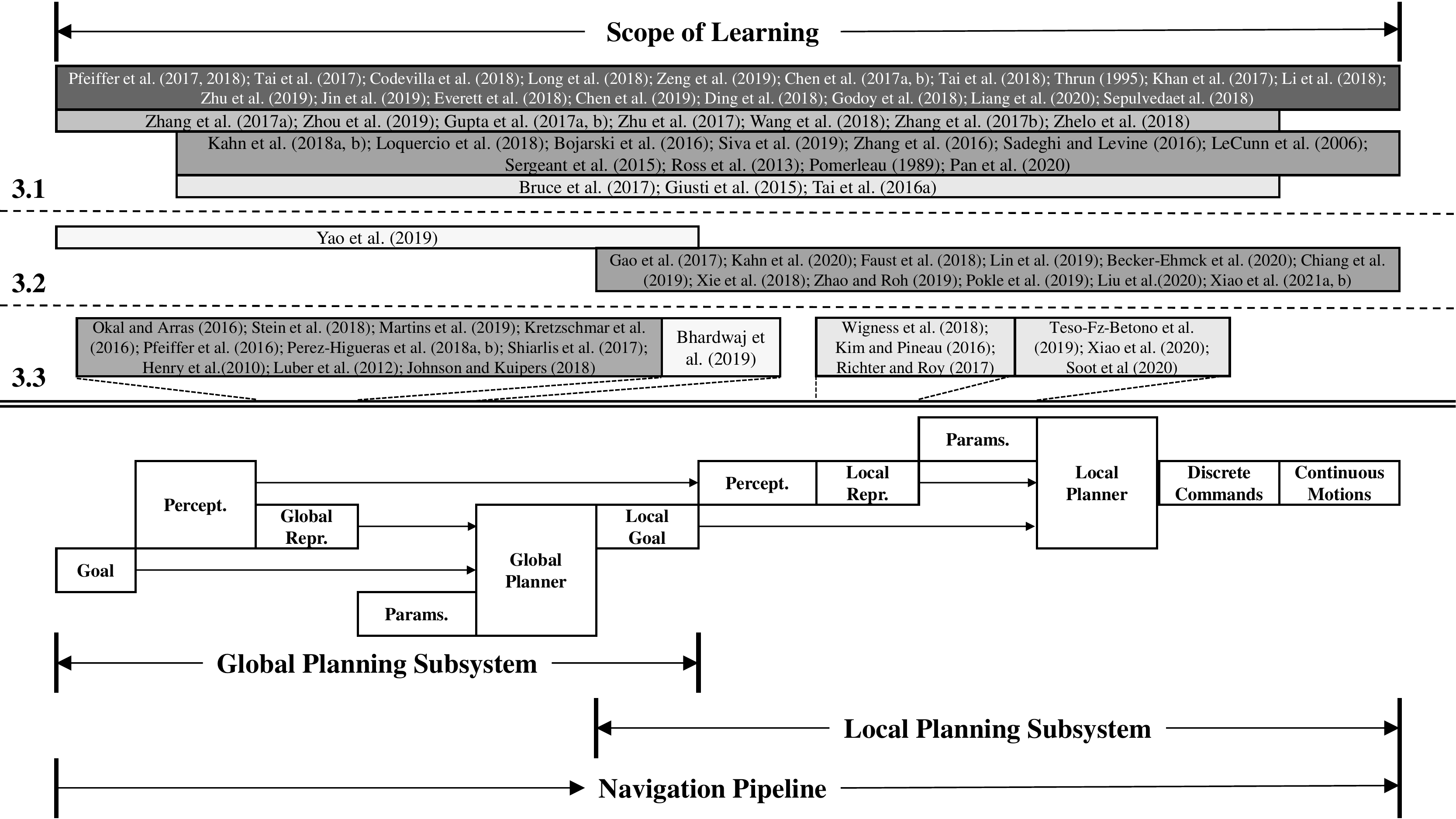}}
\end{figure*}

\begin{figure*}
\centering
\caption{Structure of Section \ref{sec::learning}}
\label{tab::learning_scope}
\includegraphics[width=1\textwidth]{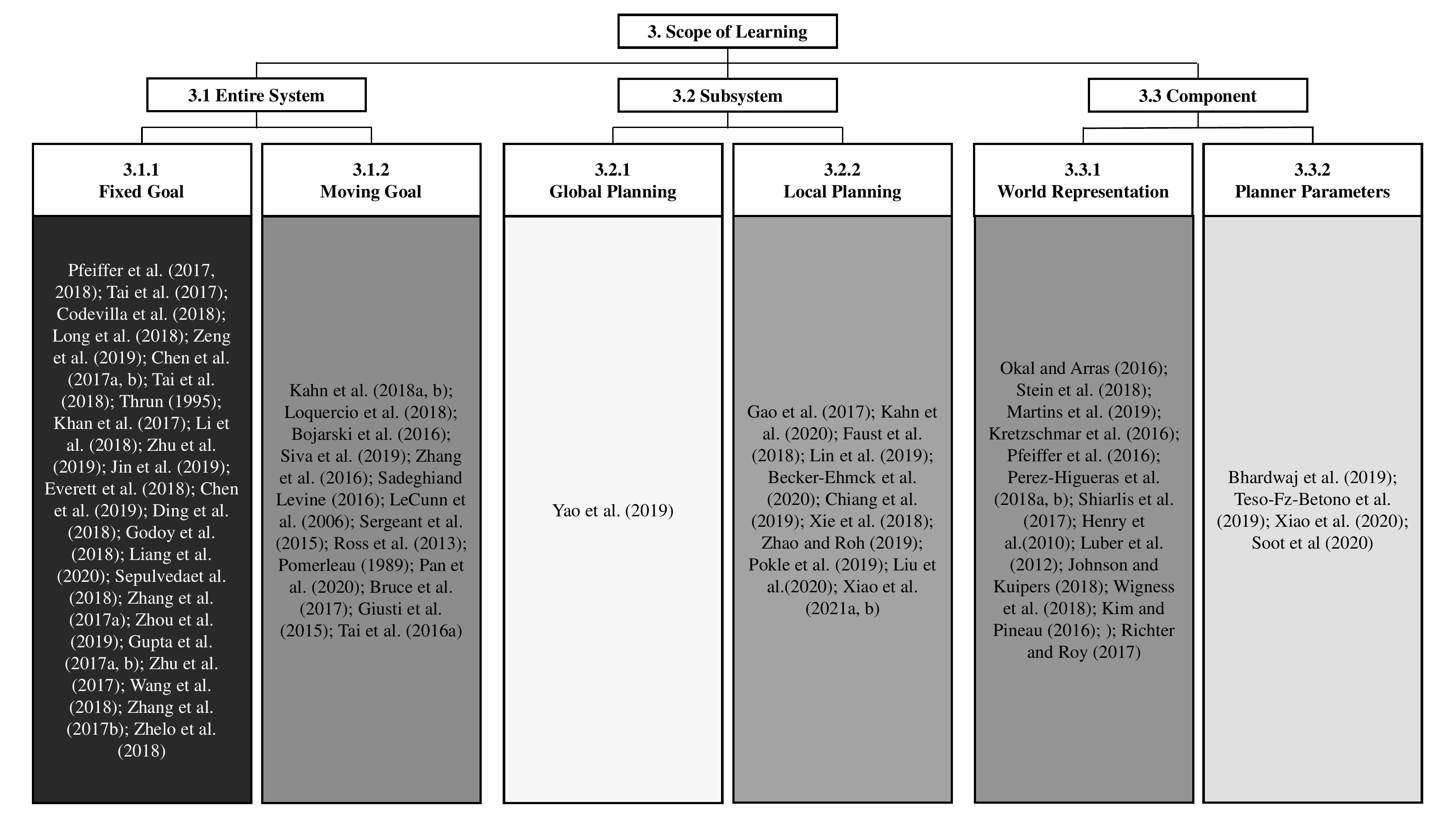}
\end{figure*}

Due to the popularity of end-to-end learning, a large body of work has been proposed to approach the navigation problem in an end-to-end fashion. That is, given raw perceptual information, these approaches seek to learn how to directly produce motion commands to move the robot either towards a pre-specified goal, or just constantly forward without any explicit intermediate processing steps. In contrast, other work in the literature has proposed to ``unwrap" the navigation pipeline and use machine learning approaches to augment or replace particular navigation subsystems or components. Therefore, we divide Section \ref{sec::learning} into three major subsections: 
\begin{enumerate}
    \item learning to replace the entire navigation stack (Section \ref{sec::learning_entire}),
    \item learning navigation subsystems (Section \ref{sec::learning_subsystem}), and
    \item learning navigation components within the navigation stack (Section \ref{sec::learning_component}). 
\end{enumerate}
A more detailed breakdown of the scope of learning targeted by each of the surveyed papers is shown in Figure \ref{tab::nav_table}. The upper portion of Figure \ref{tab::nav_table} contains works that seek to completely replace the classical navigation stack, with differences such as fixed (global) goal or moving (local) goal, and discrete or continuous motion commands. The two middle portions of the table pertain to works that seek to use learning for particular navigation subsystems --- including learning for global or local planning --- and learning methods for individual navigation components. The shade of each cell corresponds to the number of surveyed papers in each category (darker means more). The lower portion shows the corresponding components of the classical navigation pipeline. The structure of Section \ref{sec::learning} is illustrated in Figure \ref{tab::learning_scope}.

\vspace{10pt}

\subsection{Learning the Entire Navigation Stack}
\label{sec::learning_entire}
Due to the recent popularity of end-to-end machine learning techniques, work in this category comprises the majority of attempts to use machine learning to enable mobile robot navigation. Tackling the entire navigation problem in an end-to-end fashion is relatively straightforward: there is no need to formulate specific subcomponents of the classical system or relationships between these subcomponents. Instead, most works here treat the entire system as a black box with raw or processed perceptual signals as input and motion commands as output (either high-level discrete commands or low-level continuous commands). Moreover, whereas classical navigation systems always take as input a pre-specified, fixed goal (destination), we have found --- surprisingly --- that a very large number of learning-based navigation methods in this category do not use any such information. Rather, these methods merely react to the current environment around the robot, e.g., navigating the robot forward as if there were a moving goal in front of it. Therefore, we divide Section \ref{sec::learning_entire} into fixed-goal (Section \ref{sec::fixed_goal}, the first two boxes of Figure \ref{tab::nav_table}) and moving-goal navigation (Section \ref{sec::moving_goal}, the third and forth boxes of Figure \ref{tab::nav_table}).

\vspace{8pt}

\subsubsection{Fixed-Goal Navigation}
\label{sec::fixed_goal}
One of the earliest attempts at using end-to-end machine learning for fixed-goal navigation is Thrun's 1995 work \cite{thrun1995approach} --- an early proof-of-concept for completely replacing the classical sense-plan-act architecture with a single learned policy. He used Q-learning~\cite{watkins1992q} to find a policy that mapped visual, ultrasonic, and laser state information directly to discrete motion actions for the task of servoing to a designated target object. The generated motion trajectories were relatively simple, and the system sought only to move the platform toward the target rather than avoid obstacles.

Since Thrun's seminal work above, the machine learning community has proposed several other end-to-end approaches for fixed-goal navigation. Below, we organize them according to the sensory inputs presented to the learner: {\em Geometric navigation} techniques use sensory inputs that directly indicate where obstacles and free space lie in the environment (e.g., LiDAR sensors); {\em non-geometric navigation} techniques are designed for sensors that do not directly provide such information (e.g., RGB cameras); and {\em hybrid navigation} techniques utilize a combination of the two.

\vspace{6pt}

\paragraph{Geometric Navigation}
Given the ubiquity of geometric sensors on existing robotic platforms, several end-to-end machine learning techniques for fixed-goal, geometric navigation have been proposed. Many such methods propose replacing the entire navigation system with a deep neural network.

In the single-agent setting, Pfeiffer, et al. \cite{pfeiffer2017perception} presented a representative technique of those in this category, i.e., they enabled fixed-goal navigation with collision avoidance using an end-to-end neural network that maps raw LiDAR returns and a fixed goal location to low-level velocity commands. They trained the network using \gls{il} \cite{russell2016artificial} with the objective to mimic the classical \gls{ros} \texttt{move\textunderscore base} navigation stack that uses a global (Dijkstra's) and local (\gls{dwa} \cite{fox1997dynamic}) planner. Tai, et al. \cite{tai2017virtual} showed that such systems could achieve these same results even with very low-dimensional LiDAR data (10 returns) when trained using \gls{rl} \cite{russell2016artificial}, where the robot learns via trial-and-error. To facilitate better adaptation to new situations (e.g., changing navigation goals and environments) without the classical localization, mapping, or planning subsystems, Zhang, et al. \cite{zhang2017deep} incorporated \gls{drl} and successor features. With a limited, discrete action space (stay, left, right, and forward), their results showed that Deep \gls{cnn} could indeed handle new situations in both simulated and physical experiments. Zeng, et al. \cite{zeng2019navigation} instead considered a setting with dynamic obstacles, and found success using the \gls{a3c} algorithm ~\cite{mnih2016asynchronous} augmented with additional strategies such as reward shaping, the inclusion of \gls{gru} enabled memory~\cite{chung2014empirical}, and curriculum learning. Zhelo, et al. \cite{zhelo2018curiosity} also used \gls{rl} to train their system and show that adding an intrinsic reward can lead to faster learning. Zhang, et al. \cite{zhang2017neural} took these augmentations even further, and proposed neural \gls{slam} to better enable exploration behaviors (as opposed to goal-seeking beahviors) by using a neural network that contained explicit, differentiable memory structures for learning map representations. The network was trained using \gls{a3c} with an exploration-encouraging reward function. A similar idea was explored by Khan, et al. \cite{khan2017memory}, in which a \gls{macn} was proposed to enable navigation performance on par with more classical A* algorithms in partially-observable environments where exploration is necessary. Another end-to-end approach in this category is the work by Wang, et al. \cite{wang2018learning}, in which a modular \gls{drl} approach was adopted, i.e. one \gls{dqn} for global and another two-stream (spatial and temporal) \gls{dqn} for local planning. Interestingly, despite all the success noted above, Pfeiffer, et al. \cite{pfeiffer2018reinforced} performed an extensive case study of such end-to-end architectures, and ultimately concluded that such systems should not be used to completely replace a classical navigation system, particularly noting that a classical, map-based, global path planner should be used if possible, in order to reliably navigate in unseen environments. 

End-to-end learning techniques for fixed-goal, geometric navigation have also been proposed for navigation systems designed to operate in multi-agent settings.
For example, both Chen, et al. \cite{chen2017decentralized} and Long, et al. \cite{long2018towards} have enabled collision avoidance in multi-robot scenarios using \gls{drl} to train networks that map LiDAR information and a goal location directly to low-level navigation commands.
Both Tai, et al. \cite{tai2018socially} and Jin, et al. \cite{jin2019mapless} have shown that systems similar to the ones above can also be used to enable navigation in the presence of human pedestrians using \gls{rl} and geometric sensor inputs.
Ding, et al. \cite{ding2018hierarchical} showed that a similar capability can also be achieved using \gls{rl} to train a system to choose between target pursuit and collision avoidance by incorporating a \gls{hmm} \cite{stratonovich1965conditional} into a hierarchical model.
Considering the specific case in which some humans in the scene can be assumed to be companions, Li, et al. \cite{li2018role} showed that end-to-end learned approaches could also enable \gls{scn}, i.e., navigation in which the robot not only needs to avoid collisions as in previous work, but also needs to maintain a sense of affinity with respect to the motion of its companion.
Recently, Liang, et al. \cite{liang2020crowdsteer} used multiple geometric sensors (2D LiDAR and depth camera) and \gls{rl} to train an end-to-end collision avoidance policy in dense crowds. 

\vspace{6pt}

\paragraph{Non-geometric Navigation} In contrast to the approaches above that take geometric information as input, researchers have also proposed several learned, end-to-end, fixed-goal navigation systems that exclusively utilize non-geometric sensor information such as RGB imagery.
For example, Codevilla, et al. \cite{codevilla2018end} considered the problem of road navigation for a single autonomous vehicle using RGB images and a high-level goal direction (rather than an explicit goal location), and enabled such navigation using an end-to-end deep \gls{il} approach. Their system was evaluated experimentally both in simulation and in the real world. 
Zhu, et al. \cite{zhu2017target} considered a slightly different problem in which they sought to enable collision-free navigation from RGB images but where the fixed goal is specified as the image the robot is expected to see when it reaches the desired pose. They showed that an end-to-end deep \gls{rl} approach could successfully train a network that mapped camera images directly to discrete local waypoints for a lower-level controller to follow. 
A similar indoor visual navigation problem has been addressed by Sepulveda, et al. \cite{sepulveda2018deep}, in which they used imitation learning with \gls{rrt}* \cite{karaman2011sampling} as expert and utilized indoor semantic structures to produce motor controls. 
Gupta, et al. \cite{gupta2017cognitive} and Gupta, et al. \cite{gupta2017unifying} also sought to enable this style of navigation, but more explicitly looked at incorporating more classical systems into the neural network -- they jointly trained a neural-network-based mapper and planner using the \gls{vin} \cite{tamar2016value} technique. This approach resulted in the network generating a latent map representation from which the learned planner was able to generate discrete actions (stay, left, right, and forward) that enabled successful navigation to a fixed goal.

\vspace{6pt}

\paragraph{Hybrid Navigation} Researchers studying end-to-end machine learning for fixed-goal navigation have also proposed systems that combine both geometric and non-geometric sensor input. For example, for the problem of single-agent navigation with collision avoidance, Zhou, et al. \cite{zhou2019towards}
modified the traditional paradigm of hierarchical global and local planning by proposing a new switching mechanism between them. In a normal operating mode, a global planner trained using \gls{bc} used a visual image and goal as input and output one of three discrete (left, straight, and right) actions. However, when an obstacle was detected using LiDAR, the system switched to use a local planner (which can output more fine-grained actions) that was trained using \gls{rl}. 
In a continuation of their previous CADRL work \cite{chen2017decentralized}, Chen, et al. \cite{chen2017socially} also proposed a hybrid approach called \gls{sacadrl}, in which \gls{drl} over both visual and LiDAR inputs is used to train a navigation system that learns social norms like left- and right-handed passing.
Further along that same line of research, Everett, et al. \cite{everett2018motion} used both a 2D LiDAR and three RGB-D cameras in their proposed GA3C-CADRL technique, which relaxes assumptions about the other agents' dynamics and allows reasoning about an arbitrary number of nearby agents.

\vspace{6pt}

\paragraph{Exteroception}
In contrast to the previous sections which considered sensor input collected using sensors on the robot itself, there also exists some end-to-end learned, fixed-goal navigation work that instead uses {\em exteroception}, or perception data gathered using sensors external to the robot. For example, in the robot soccer domain, using low-level state information estimated from overhead cameras, Zhu, et al. \cite{zhu2019learning} learned a policy trained using \gls{rl} that maps the state information directly to motion commands to enable several goal-oriented motion primitives, including \texttt{go-to-ball}, in which a single small-size robot navigates to the ball without any obstacle or other players on the field.
This work by Zhu, et al. \cite{zhu2019learning} is similar to Thrun's work mentioned previously \cite{thrun1995approach}, where the robot also learned to pursue a target, but used \gls{ddpg} \cite{lillicrap2015continuous} and an overhead camera instead of Q-learning and onboard perception, respectively. 
Godoy, et al. \cite{godoy2018alan} trained a planner with \gls{rl} and global tracking in simulation and modified the learned velocities with \gls{orca} \cite{van2011reciprocal} for multi-robot navigation. 
Chen, et al. \cite{chen2019crowd} also proposed a method for social navigation that used exteroceptive ground-truth simulation data and \gls{drl} to model human-robot and human-human interactions for motion planning. 

\vspace{6pt}

The works described in Section \ref{sec::fixed_goal} all sought to replace the entire navigation pipeline with a single, end-to-end learned function. The approaches use as input both a fixed goal location
and sensor data (geometric, non-geometric, or both), and produce as output discrete or continuous motion commands to drive the robot to the goal location. Results from this body of work show that such learned navigation systems can enable obstacle avoidance behavior, work in the presence of other robots, consider human social norms, and explore unknown environments.

\vspace{8pt}

\subsubsection{Moving-Goal Navigation}
\label{sec::moving_goal}
While moving to a specified destination is the ultimate objective of classical navigation systems, a large body of learning-based navigation approaches are designed to address a more simplified navigation problem, i.e., that of navigating the robot towards a moving goal that is always in front of the robot. Solutions to this type of problem can lead to behaviors such as lane keeping, obstacle avoidance, and terrain adaptation, but they cannot ensure that the robot arrives at a specified goal. We conjecture that the reason that this problem has gained so much traction in the learning community is that the limited scope of such problems helps constrain the learning task, which, in turn, leads to greater success with existing machine learning methods.

Early work in this category focused on creating autonomous ground vehicles, and one of the earliest such approaches was ALVINN, an \gls{alvinn} \cite{pomerleau1989alvinn}, proposed in 1989. \gls{alvinn} was trained in a supervised manner from simulated data, and during deployment, it used both image and laser inputs to produce an output direction for the vehicle to travel. The vehicle did not navigate to any specified goal but instead learned to keep moving forward down the middle of the road.
Almost two decades later, DARPA’s \gls{lagr} program further stimulated related research. LeCunn, et al. \cite{muller2006off} presented a similar work, in which the navigation system was trained in an end-to-end fashion to map raw input binocular images to vehicle steering angles for high-speed off-road obstacle avoidance using data collected from a human driver and a six-layer \gls{cnn}.
More recently, with the help of much deeper \glspl{cnn}, far more data from human drivers, and more powerful computational hardware, Bojarski, et al. \cite{bojarski2016end} built the DAVE-2 system for autonomous driving. DAVE-2 demonstrated that \glspl{cnn} can learn lane- and road-following tasks without manual decomposition into separate perception, reasoning, and planning steps.

There are also several works that have looked at enabling various types of indoor robot navigation. 
For example, Sergeant, et al. \cite{sergeant2015multimodal} proposed a method that allowed a ground robot to navigate in a corridor without colliding with the walls using human demonstration data and multimodal deep autoencoders.
Similarly, Tai, et al. \cite{tai2016deep} also proposed to use human demonstrations to learn obstacle avoidance, where they formulated the problem as one of mapping images directly to discrete motion commands.
Kahn, et al. \cite{kahn2018self} also sought to enable collision-free hallway navigation, and proposed a method that learns to produce steering angle while maintaining 2m/s speed from training experience that included sensed collisions. 
Kahn, et al. \cite{kahn2018composable} further collected event cues during off-policy training and used these event cues to enable different navigation tasks during deployment, such as avoid collisions, follow headings, and reach doors. 

In the space of outdoor navigation, Siva, et al. \cite{siva2019robot} leveraged human teleoperation and combined representation learning and imitation learning together to generate appropriate motion commands from vision input when the robot was traversing different terrain types, such as concrete, grass, mud, pebbles, and rocks. 
Pan, et al. \cite{pan2020imitation} looked at enabling agile off-road autonomous driving with only low-cost onboard sensors and continuous steering and throttle commands. Instead of human demonstration, they developed a sophisticated \gls{mpc} with expensive IMU and GPS and used it as an expert to generate training data for batch and online \gls{il} with \gls{dagger} \cite{ross2011reduction}. 

Similar navigation without a specific goal has also been applied to \glspl{uav}, for example, using \gls{il} to learn from an expert: As an early application of \gls{dagger}, Ross, et al. \cite{ross2013learning} used a human pilot's demonstration to train a \gls{uav} to fly in real natural forest environments with continuous \gls{uav} heading commands based on monocular vision. 
Aiming to navigate forest trails, Giusti, et al. \cite{giusti2015machine} used data collected from three GoPros mounted on a human hiker's head facing left, straight, and right to learn a neural network to classify a quadrotor's discrete turning right, keeping straight, and turning left actions, respectively. 
Similarly, Loquercio, et al. \cite{loquercio2018dronet} used labeled driving data from cars and bicycles to train DroNet, a \gls{cnn} that can safely fly a drone through the streets of a city. 
To avoid labeled training data, Sadeghi and Levine \cite{sadeghi2016cad2rl} looked at enabling collision-free \gls{uav} navigation in indoor environments with CAD\textsuperscript{2}RL, in which a collision-avoidance \gls{rl} policy was represented by a deep \gls{cnn} that directly processes raw monocular images and outputs velocity commands. 
Similar to the idea of using a sophisticated \gls{mpc} as an expert \cite{pan2020imitation}, Zhang, et al. \cite{zhang2016learning} also used \gls{mpc} with full access to simulated state information for \gls{rl} training. During their simulated deployment, only observation (e.g., onboard LiDAR) was available, in contrast to full state information. 

One special case is by Bruce, et al. \cite{bruce2017one}, in which the training environment was built with a 360\degree~camera and the robot was trained to only navigate to one fixed goal specified during training. Similar to the moving-goal approaches discussed above, this work does not take in a goal as input. This end-to-end approach is robust to slight changes during deployment, such as the same hallway on a different day, with different lighting conditions, or furniture layout. 

\vspace{6pt}

While the works described in Section \ref{sec::moving_goal} treated the entire navigation pipeline as an end-to-end black box, they are not applicable to the problem of navigation to a specific goal. Rather, they look to enable navigation behaviors such as lane keeping and forward movement with collision avoidance. Learning such behaviors is very straightforward and simple for proofs-of-concept of different learning methods, but its applicability to real navigation problems is limited compared to fixed-goal navigation.

\vspace{10pt}

\subsection{Learning Navigation Subsystems}
\label{sec::learning_subsystem}
Despite the popularity of end-to-end learning to replace the entire navigation system, there also exist learning for navigation approaches that instead target subsystems of the more classical architecture. More limited in scope than fully end-to-end navigation, the benefit of this paradigm is that the advantages of the other navigation subsystems can still be maintained, while the learning research can target more specific subproblems.
Most learning approaches in this category have focused on the local planning subsystem (Section \ref{sec::local_planning}), and one work has looked at global planning (Section \ref{sec::global_planning}).
Each interfaces with the other subsystem, which is implemented according to a classical approach (the fifth and sixth box of Figure \ref{tab::nav_table}).

\vspace{8pt}

\subsubsection{Learning Global Planning}
\label{sec::global_planning}
The only approach that we are aware of in the literature that seeks to use machine learning to replace the global planning subsystem is that of Yao, et al. \cite{yao2019following}. They considered the problem of building a navigation system that could observe social rules when navigating in densely-populated environment, and  
used a deep neural network as a global planner, called Group-Navi GAN, to track social groups and generate motion plans that allowed the robot to join the flow of a social group by providing a local goal to the local planner. Other components of the existing navigation pipeline, such as state estimation and collision avoidance, continued functioning as usual.

\vspace{8pt}

\subsubsection{Learning Local Planning}
\label{sec::local_planning}
Compared to the global planner, the literature has focused much more on replacing the local planner subsystem with machine learning modules. These works use the output from classical global planners to compute a local goal, combine that goal with current perceptual information, and then use learned modules to produce continuous motion control. In the following, we discuss learning approaches in ground, aerial, and marine navigation domains.

For ground navigation, Intention-Net \cite{gao2017intention} was proposed to replace the local planning subsystem, where continuous speed and steering angle commands were computed using ``intention'' (local goal) from a global planner (A*) in the form of an image describing the recently traversed path, along with other perceptual information. 
The system was trained end-to-end via \gls{il}. 
A similar approach was taken by Faust, et al. \cite{faust2018prm} in their PRM-RL approach: they used \gls{prm} as a global planner, which planned over long-range navigation tasks and provided local goals to the local planner, which was an \gls{rl} agent that learned short-range point-to-point navigation policies. The \gls{rl} policy combined the local goal with LiDAR perception and generated continuous motion commands. 
In their further PRM-AutoRL work \cite{chiang2019learning}, they used AutoRL, an evolutionary automation layer around \gls{rl}, to search for a \gls{drl} reward and neural network architecture with large-scale hyper-parameter optimization. 
Xie, et al. \cite{xie2018learning} instead investigated the idea of using a classical PID controller to  ``kick-start'' \gls{rl} training to speed up the training process of the local controller, which took local goal as input from a classical global planner. 
Pokle, et al. \cite{pokle2019deep} targeted at local planning to address social navigation: with the help of a classical global planner, the learned local planner adjusts the behavior of the robot through attention mechanisms such that it moves towards
the goal, avoids obstacles, and respects the space of nearby
pedestrians. 
Recently, Kahn, et al. \cite{kahn2020badgr} introduced BADGR to plan actions based on non-geometric features learned from physical interaction, such as collision with obstacles, traversal over tall grass, and bumpiness through uneven terrains.  Although this method did not require simulations or human supervision, the robot experienced catastrophic failure during training, e.g., flipping over, which required manual intervention and could have damaged the robot. 
Xiao, et al. \cite{xiao2020toward} proposed \gls{lfh} (along with a Sober Deployment~\cite{xiao2020agile} and a learned hallucination extension~\cite{wang2021from}), a paradigm that safely learns local planners in free space and can perform agile maneuvers in highly-constrained obstacle-occupied environments. 
Liu, et al. \cite{liu2020lifelong} introduced Lifelong Navigation, in which a separate neural planner is learned from self-supervised data generated by the classical \gls{dwa} planner. During deployment, the learned planner only complements the \gls{dwa} planner, when the latter suffers from suboptimal navigation behaviors. The neural planner trained using \gls{gem} \cite{lopez2017gradient} can avoid catastrophic forgetting after seeing new environments. 
More recently, Xiao, et al. \cite{xiao2021learning} utilized inertial sensing to embed terrain signatures in a continuous manner and used learning to capture elusive wheel-terrain interactions to enable accurate, high-speed, off-road navigation.

For aerial navigation, Lin, et al. \cite{lin2019flying} used \gls{il} to learn a local planning policy for a \gls{uav} to fly through a narrow gap, and then used \gls{rl} to improve upon the achieved results. 
Becker, et al. \cite{becker2020learning} developed the first \gls{mbrl} approach deployed on a real drone where both the model and controller were learned by deep learning methods. 
This approach can achieve point-to-point flight with the assistance of an external \gls{mocap} system and without the existence of obstacles. 

Moving to marine navigation, Zhao and Roh \cite{zhao2019colregs} used \gls{drl} for multiship collision avoidance: a \gls{drl} policy directly mapped the fully observable states of surrounding ships to three discrete rudder angle steering commands. A unified reward function was designed for simultaneous path following (from global planner) and collision avoidance. 

The works described in Section \ref{sec::local_planning} addressed only the local planning part of the traditional navigation stack. These works replaced the classical local planner with learning approaches, which take as input a local goal from the global planner along with the current perception. 

\vspace{10pt}

\subsection{Learning Individual Components}
\label{sec::learning_component}
Maintaining the traditional structure of classical navigation approaches, a relatively small body of literature has used machine learning techniques to learn individual components (the seventh box of Figure \ref{tab::nav_table}) rather than the entire navigation system or entire subsystems. Section \ref{sec::learning_component} focuses on these works, including improving world representation (Section \ref{sec::world_representation}) and fine-tuning planner parameters (Section \ref{sec::planner_parameters}). 

\vspace{8pt}

\subsubsection{World Representation}
\label{sec::world_representation}
One popular component of a classical navigation system that the machine learning community has targeted is that of world representation, which, in the classical navigation stack, acts as a bridge between the perceptual streams and the planning systems.

For social navigation problems, Kim and Pineau \cite{kim2016socially} used \gls{irl} to infer cost functions: features were first extracted from RGB-D sensor data, and a local cost function over these features was learned from a set of demonstration trajectories by an expert using \gls{irl}. The system operated within the classical navigation pipeline, with a global path planned using a shortest-path algorithm, and local path using the learned cost function to respect social constraints. 
Johnson and Kuipers \cite{johnson2018socially} collected data using a classical planner \cite{park2016graceful} and observed human navigation patterns on a campus. Social norms were then learned as probability distribution of robot states conditioned on observation and action, before being used as an additional cost to the classical obstacle cost. 
Other researchers tackled cost function representation at a more global level. 
Luber, et al. \cite{luber2012socially} utilized publicly available surveillance data to extract human motion prototypes and then learn a cost map for a \textit{Theta}* planner \cite{daniel2010theta}, an any-angle A* variant to produce more natural and shorter any-angle paths than A*. 
Henry, et al. \cite{henry2010learning} used \gls{irl} for cost function representation and enabled behaviors such as moving with the flow, avoiding high-density areas, 
preferring walking on the right/left side, and reaching the goal quickly. 
Okal and Arras \cite{okal2016learning} developed a graph structure and used Bayesian \gls{irl} \cite{ramachandran2007bayesian} to learn the cost for this representation. With the learned global representation, traditional global planner (A*) planned a global path over this graph, and the POSQ steering function \cite{palmieri2014efficient} for differential-drive mobile robots served as a local planner. 
A similar approach, ClusterNav, was taken by Martins, et al. \cite{martins2019clusternav}, which learns a socially acceptable global representation and then also uses A* for global path planning.
Using \gls{rrt} as global planner, Shiarlis, et al. \cite{shiarlis2017rapidly} and Perez, et al. \cite{perez2018teaching} also used \gls{irl} to learn its cost function for social navigation. 
Perez, et al. \cite{perez2018learning} learned a path predictor using fully convolutional neural networks from demonstrations and used the predicted path as a rough costmap to bias the \gls{rrt} planner. 
Kretzschmar, et al. \cite{kretzschmar2016socially} used maximum entropy probability distribution to model social agents' trajectory with geometric input and then plan based on that information. 
Similar to this maximum entropy probability distribution model \cite{kretzschmar2016socially}, the work by Pfeiffer, et al. \cite{pfeiffer2016predicting} used RGB input.

Learned world representations have also been used for other styles of navigation beyond social compliance.
In particular, Wigness, et al. \cite{wigness2018robot} used human demonstration and maximum entropy \gls{irl} \cite{ziebart2008maximum} to learn a local costmap so that the robot can mimic a human demonstrator's navigation style, e.g., maintaining close proximity to grass but only traversing road terrain, ``covert'' traversal to keep close to building edges and out of more visible, open areas, etc.
Richter and Roy \cite{richter2017safe} used antoencoder to classify novelty of the current visual perception compared to the body of existing training data: if the visual perception is not novel, a learned model predicts collision. Otherwise, a prior estimate is used. The predicted collision serves as part of the cost function for planning. 
Finally, for navigating to a global goal in structured but unknown environments, Stein, et al. \cite{stein2018learning} defined subgoals to be on the boundary between known and unknown space, and used neural networks to learn the cost-to-go starting from a particular subgoal of the global goal, such as the probability of getting stuck in a dead-end after going to that subgoal. 

\vspace{8pt}

\subsubsection{Planner Parameters}
\label{sec::planner_parameters}
Another way in which machine learning approaches have targeted individual components of the classical navigation pipeline is by tuning the parameters of the existing systems.
For example, classical planners have a set of tunable parameters which are designed for the planner to face different situations or planning preferences, e.g., inflation radius, sampling rate, and trajectory optimization weights. A relatively small amount of machine learning research has looked at using learning techniques to achieve parameter fine-tuning, instead of using extensive human expertise and labor.

At the global planner level, Bhardwaj, et al. \cite{bhardwaj2019differentiable} proposed a differentiable extension to the already differentiable \gls{gpmp2} algorithm, so that a parameter (obstacle covariance) could be learned from expert demonstrations. Through backpropagation, \gls{gpmp2} with the learned parameter can find a similar global path to the expert's. 

Similar ideas have been applied at the local planning level. Teso, et al. \cite{teso2019predictive} proposed Predictive \gls{dwa} by adding a prediction window to traditional \gls{dwa} \cite{fox1997dynamic}, and used an \gls{anfis} to optimize each of the fixed parameters' values, i.e. \gls{dwa}'s optimization weights, to increase performance. Through hand-engineered features and backpropagation, the three weights could be modified individually by three \gls{anfis}'s. 

\begin{figure}
\centering
\includegraphics[width=1\columnwidth]{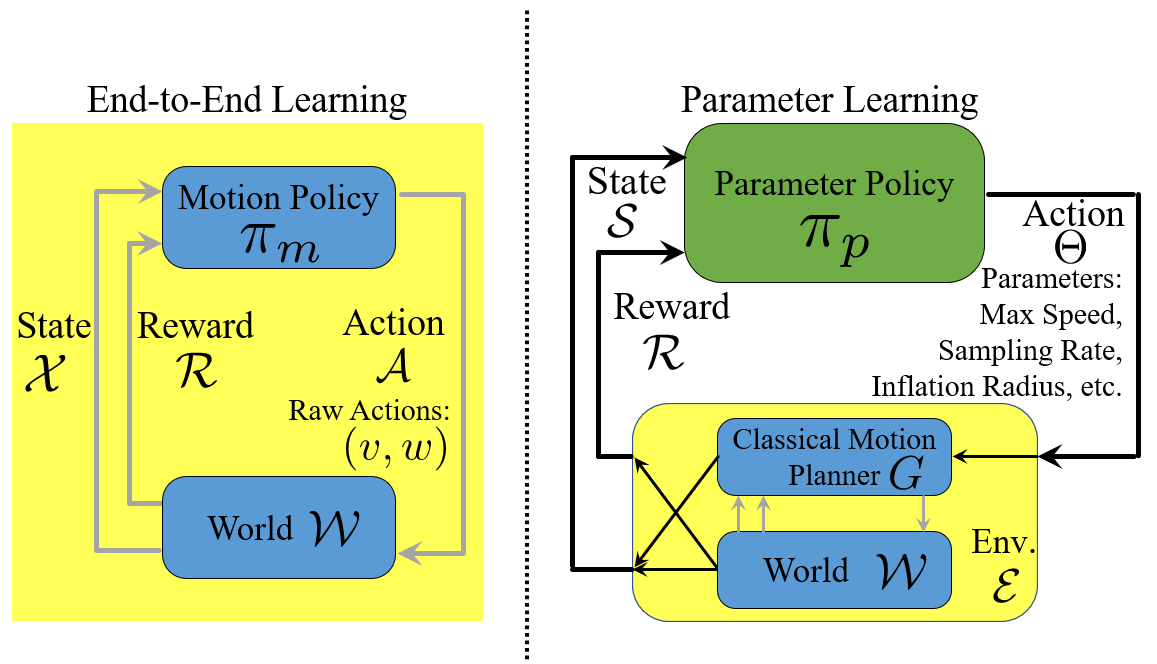}
\caption{Learning End-to-End Motion Policy vs. Learning Parameter Policy (reproduced with authors' permission \cite{xiao2021appl}).}
\label{fig::paradigms}
\end{figure}

More recently, Xiao, et al. \cite{xiao2021appl} proposed \gls{appl} from different human interaction modalities, including teleoperated demonstration (APPLD \cite{xiao2020appld}), corrective interventions (APPLI \cite{wang2020appli}), evaluative feedback (APPLE \cite{wang2021apple}), and reinforcement learning (APPLR \cite{xu2020applr}). \gls{appl} introduces the parameter learning paradigm, where the learned policy does not directly issue end-to-end motion commands to move the robot (as shown in Figure \ref{fig::paradigms} left), but interfaces with a classical motion planner through its hyper-parameters (Figure \ref{fig::paradigms} right). The learned parameter policy adjusts the planner parameters on the fly to adapt to different navigation scenarios during runtime and outperforms planners with fixed parameters fine-tuned by human experts. 

A similar work to learning adaptive planner parameters is learning adaptive motion primitives \cite{sood2020learning}. Although not implemented on any physical mobile robots, it is shown that on a 3 degree-of-freedom motion planning problem for navigation using the Reeds-Shepp path, the learned adaptive motion primitives in conjunction with search algorithms lead to
over 2x speedup in planning time.

\vspace{10pt}

The works described in Sections \ref{sec::learning_subsystem} and \ref{sec::learning_component} maintained the classical navigation pipeline and many existing components. The learning approaches proposed only aimed at the global planner, local planner, or individual components, such as improving world representation and fine-tuning planner parameters. The most important benefit of these methods is that they maintain the advantages of the classical approaches, e.g., safety, explainability, and reliability, and improve upon their weaknesses simultaneously.

\vspace{10pt}

\subsection{Analysis}

In this section, we reviewed learning approaches that replace (1) the entire navigation stack, (2) navigation subsystems, and (3) individual components within the navigation stack. The scope of learning of each work is illustrated at the top of Figure \ref{tab::nav_table}, with their corresponding navigation components at the bottom. For end-to-end approaches, we reviewed works that are goal-oriented (first two boxes of Figure \ref{tab::nav_table}) and works that do not consider the navigation goal (third and forth box). In contrast to end-to-end approaches, we discussed subsystem-level learning, i.e. global (fifth box) and local (sixth box) planning, and also learning individual components (seventh box). 

Considering all of these works together, we make the following observations:
\begin{enumerate}
\item \textbf{The majority of machine learning for navigation work focuses on end-to-end approaches, despite their lack of proven reliable applicability in real-world scenarios.} In particular, 43 out of the 74 surveyed papers focus on end-to-end techniques as illustrated in detail in Figures \ref{tab::nav_table} and \ref{tab::learning_scope}. The functional blocks within the classical navigation pipeline are obscured by the end-to-end learning approaches, which directly map from sensory input to motion. An important benefit of the end-to-end paradigm is that it is very straightforward and requires relatively little robotics engineering effort. Roboticists no longer need to hand-craft the navigation components and hand-tune their parameters to adapt to different platforms and use cases. 
Another appeal of end-to-end methods is that they have the potential to avoid the prospect of cascading errors such as errors in the global representation being propagated to the local planning level. 
With end-to-end learning, every component within the entire navigation stack is connected and learned jointly. The cascading errors will be eliminated and only one end-to-end error will be minimized by data-driven approaches. However, end-to-end navigation approaches have many problems as well, in addition to the intrinsic problems of learning methods in general, such as high demand on training data, overfitting, and lack of explainability. 

\item \textbf{One third of the end-to-end learning approaches lack the ability to navigate to user-defined goals.} These approaches instead operate in the context of the robot perpetually moving forward, and seek to enable behaviors such as lane keeping and obstacle avoidance. The proposed approaches are good at enabling these reactive behaviors, but they lack the interface to a pre-specified fixed global goal. The capability to navigate to a specific goal is very common in classical navigation approaches. The lack of such capability in many learning-based systems is possibly due to the drawback of learning methods: It is very difficult for a learned model to generalize to arbitrarily defined goals, which is not seen in its training data. As the two main inputs to navigation systems, perception and goal, the space of the former is usually easy to be captured by training data, while the later is difficult. For example, a robot may have learned how to avoid obstacles based on the current perception of the surroundings and navigate to a goal one meter away from it, but it may find it hard to navigate to a goal ten meters away, 100 meters away, or even 1000 meters away. Not being able to see similar goal configurations during training makes it hard to generalize to arbitrarily specified goals. As pointed out by Pfeiffer, et al. \cite{pfeiffer2018reinforced}, an end-to-end learning architecture cannot completely replace a map-based path planner in large complex environments.

\item \textbf{All of the non-end-to-end approaches surveyed can navigate to user-defined goals.} In response to the shortcomings of end-to-end learned systems, 31 out of the 74 surveyed papers used non-end-to-end approaches, and all of them can navigate to user-defined goals. For these approaches, designers can retain the desirable properties of existing navigation components while also studying the benefits of learning other components. These benefits include alleviating the difficulty of manually setting cost functions or planner parameters and also increasing the overall performance of the system.

\item \textbf{Approaches that seek to learn subsystems have focused predominately on local planning rather than global planning.} Of the 13 subsystems approaches, 12 focus on local planning, while only one considers global planning. We posit that this lack of focus on global planning may be because it is easier to generalize to a close local goal or local segment from a global path than an unseen, faraway global goal. 
    
\end{enumerate}

The analysis of the findings in this section leads us to propose a hybrid classical and learning navigation paradigm as a very promising direction for future research (Item 1 in Section \ref{sec::directions}).

\vspace{12pt}
\section{COMPARISON TO CLASSICAL APPROACHES}
\label{sec::comparison}

\newacronym{pfm}{PFM}{Potential Field Method}
\newacronym{lstm}{LSTM}{Long Short Term Memory network}
\newacronym{gail}{GAIL}{Generative Adversarial Imitation Learning}

Whereas Section \ref{sec::learning} above focused on reviewing where and how existing learning approaches fit into the context of the classical navigation pipeline, this section focuses on comparing the results of learning methods to those of classical approaches.

\begin{figure*}
\centering
\caption{Comparison to Classical Approaches with Section Headings}
\label{tab::comparison}
\includegraphics[width=\textwidth]{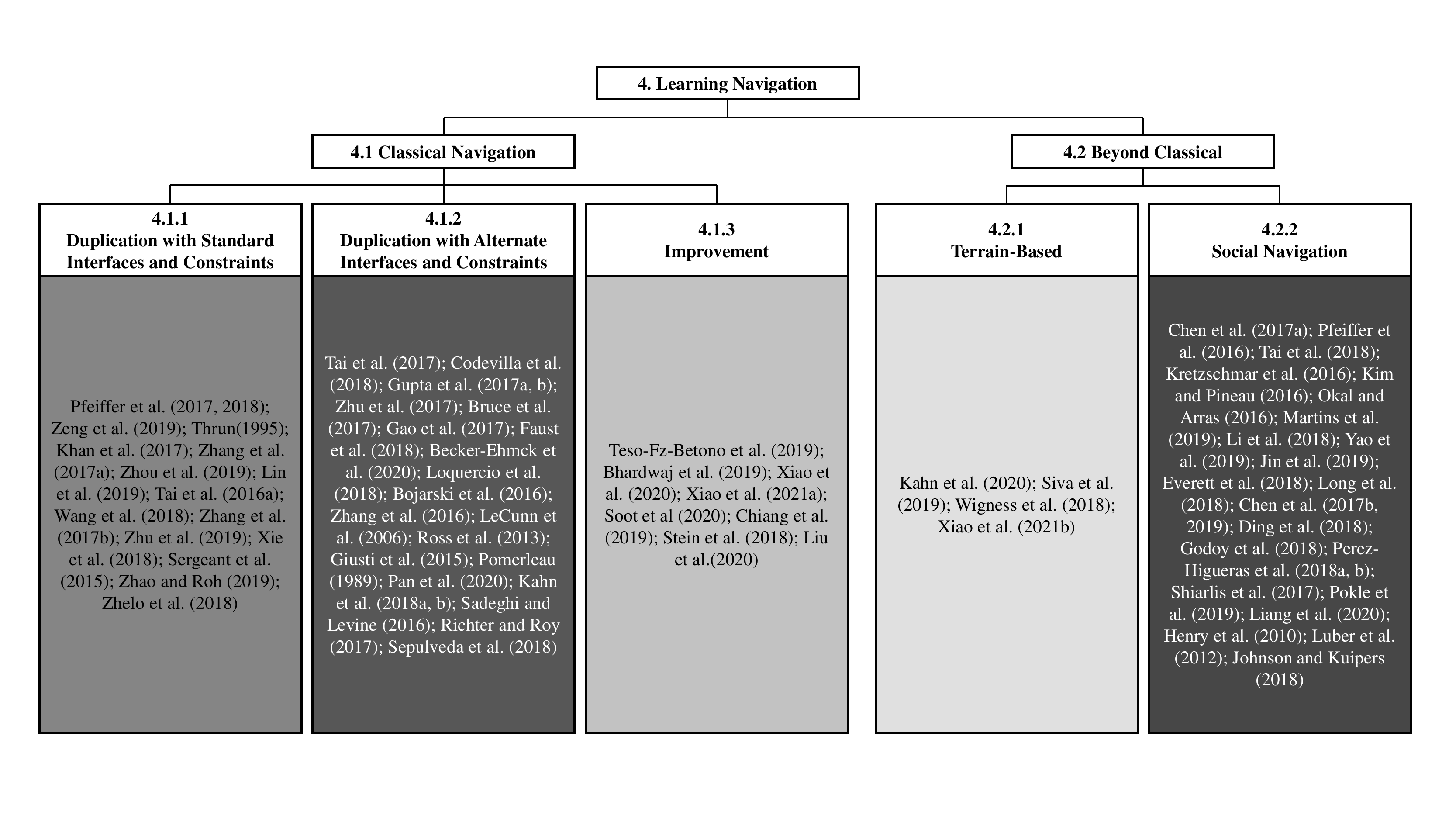}
\end{figure*}

Classical navigation systems seek to use current perceptual input to produce collision-free motion commands that will navigate the robot to a pre-specified goal. While there exist many learning-based navigation approaches that seek to solve the navigation problem in this classical sense, a portion of the literature also attempts to go beyond that problem and provide additional capabilities. Therefore, we divide the discussion of this section based on these two categories: {\em learning for navigation in the classical sense} (Section \ref{sec::learning_classical}) and {\em learning beyond classical navigation} (Section \ref{sec::beyond_classical}).
Note that the same reviewed papers in Section \ref{sec::learning} are analyzed in this section again. The comparison categories and the corresponding literature are shown in Figure \ref{tab::comparison}. 

\vspace{10pt}

\subsection{Learning for Navigation in the Classical Sense}
\label{sec::learning_classical}
The autonomous navigation problem that has been classically considered is that of perceiving the immediate environment and producing collision-free motion commands to move the robot toward a pre-specified goal. Although many learning-based approaches for navigation do not enable goal-oriented navigation (see Section \ref{sec::learning}), the key aspect of the classical navigation problem we focus on here is that it seeks to produce collision-free motion based on geometric perception, such as LiDAR range readings or depth images, and a relatively accurate motion model. In this section, we discuss learning methods that seek to enable this same behavior. Broadly speaking, we have found that the machine learning literature in this area can be divided into three categories: (1) machine learning approaches that merely aim to {\em duplicate} the input-output behavior of classical approaches with standard  interfaces and constraints (Section \ref{sec::duplication}); (2) machine learning approaches that aim to provide the same capability as classical approaches but can {\em leverage alternate interfaces and constraints} (e.g., sensor data, assumptions, and/or computational constraints) compared to standard approaches (Section \ref{sec::different_approach}); and (3) machine learning approaches that explicitly aim to provide navigation behaviors that can {\em outperform} classical approaches (Section \ref{sec::improvement}). 

\vspace{8pt}
\subsubsection{Duplication with Standard Interfaces and Constraints}
\label{sec::duplication}
An initial goal for machine learning methods for motion planning and control in mobile robot navigation in the literature has been that of duplicating the input-output behavior of classical navigation approaches with standard interfaces and constraints. Because achieving this goal would result in a system with behavior indistinguishable from that of an existing classical approach, the literature in this area is certainly of more interest to the machine learning community than to the robotics community. However, duplicating the results of decades of engineering effort by simply using learning-based approaches---especially in an end-to-end fashion---can be a strong testimony to the power of machine learning.
Thus, the work belonging to this category has mostly focused on the typical concerns of the machine learning community rather than that of the robotics community. In fact, much of the literature does not directly compare the resulting navigation performance to that achievable with classical approaches. Instead, approaches in this category are typically compared to another baseline learning approach.
Therefore, we divide work in this section into two categories: (1) initial, first-step machine learning approaches for duplicating mobile navigation, and (2) later improvements upon those initial successes, where the improvements are with respect to concerns in the learning community (e.g., amount of training data).

\vspace{6pt}

\paragraph{Initial Successes}
As first steps towards duplicating classical navigation systems, several machine learning approaches have been proposed for various versions of the navigation problem.
For example, both Thrun \cite{thrun1995approach} and Zhu, et al. \cite{zhu2019learning} proposed \gls{rl} techniques (using Q-learning and \gls{ddpg}, respectively) that aim to induce simple target pursuit without taking obstacle avoidance into account. Unfortunately, neither of these approaches compared with classical approaches.
Sergeant, et al. \cite{sergeant2015multimodal} instead considered the problem of only obstacle avoidance without seeking to achieve a specific navigation goal in mind. 
While they compared the performance of various versions of their proposed method, they also did not compare their system with any classical navigation system.
Seeking to enable moving-goal navigation with obstacle avoidance, Tai, et al. \cite{tai2016deep} proposed a technique that they claimed was the first end-to-end, model-free navigation approach that used depth images instead of LiDAR. Again, however, they only compared the navigation results to a human demonstration using supervised classification error, and did not compare navigation performance to any classical approaches. 
Pfeiffer, et al. \cite{pfeiffer2017perception} and Zhou, et al. \cite{zhou2019towards} proposed similar systems in that they are end-to-end learning approaches, but they sought to enable collision-free fixed-goal navigation using LiDAR input. They did compare their approaches to classical navigation systems: Pfeiffer, et al. \cite{pfeiffer2017perception} showed that their approach could achieve similar performance, while the approach by Zhou, et al. \cite{zhou2019towards} was always outperformed by classical navigation using \gls{slam}. 
For multiagent collision avoidance, Zhao and Roh \cite{zhao2019colregs} proposed a decentralized \gls{drl}-based approach. While they showed experimental success in a simulation environment, no comparison to any classical technique was made.

\vspace{6pt}

\paragraph{Later Improvements} 
Building upon the initial successes of applying relatively straightforward machine learning approaches to navigation discussed above, researchers have since looked into improving these various approaches in ways of particular interest to the machine learning community.

For example, building upon \gls{rl}, like the approaches by Thrun \cite{thrun1995approach} and Zhu, et al. \cite{zhu2019learning} above, 
Zhang, et al. \cite{zhang2017deep} used successor features and transfer learning in an attempt to determine whether or not \gls{cnn}-based systems could indeed duplicate the performance of classical planning systems. They showed that, with increased training epochs (8 hours of real experience), the task reward achieved by their approach outperformed \gls{dqn} and rivaled that of classical A*. However, \gls{rl} reward is possibly not an appropriate metric to evaluate navigation performance. At least in the video demonstration, the performance of \gls{rl} with successor features still looked inferior to classical approaches. 
Zeng, et al. \cite{zeng2019navigation} added reward shaping, memory \gls{gru}, and curriculum learning to \gls{a3c}, but only compared to other rudimentary \gls{a3c} versions, not classical approaches. The improvement in terms of success rate compared to \gls{a3c} was purely along the learning dimension: where it stood with respect to classical approaches is unclear. 
The results of the curiosity-driven \gls{drl} approach by Zhelo, et al. \cite{zhelo2018curiosity} showed that their \gls{drl} with intrinsic motivation from curiosity learned navigation policies more effectively and had better generalization capabilities in previously unseen environments, when compared with existing \gls{drl} algorithms, such as vanilla \gls{a3c}. They were not compared with any classical approaches.
Partially respecting the global vs. local paradigm in the classical navigation pipeline, modular \gls{rl} developed by Wang, et al. \cite{wang2018learning}, in comparison to vanilla end-to-end \gls{rl}, can achieve better results than classical \gls{pfm}. However, \gls{pfm} is far inferior to most state-of-the-art classical navigation approaches, such as those with \gls{slam}, as acknowledged by the authors. 

To address memory in navigation, Neural \gls{slam} by Zhang, et al. \cite{zhang2017neural} outperformed vanilla \gls{a3c} stacked with two \glspl{lstm}, but was not compared with classical \gls{slam} approaches. The Gazebo simulation test looked more or less like a toy example for proof-of-concept. \gls{macn} by Khan, et al. \cite{khan2017memory} used A* as an upper bound and compared the ratio of exploration path length of different learning methods against that of A*. This ratio was never smaller than one, meaning the performance was never better than the classical approach. 

Another learning-specific way in which the community has sought to improve these approaches is by seeking better and faster convergence, especially for sample-inefficient \gls{rl} approaches. One option that has been explored is to use \gls{il} to initialize the \gls{rl} policy. Following Pfeiffer, et al. \cite{pfeiffer2017perception}, the case study presented by Pfeiffer, et al. \cite{pfeiffer2018reinforced} compared the overall system performance for various combinations of expert demonstrations (\gls{il}) with \gls{rl}, different reward functions, and different \gls{rl} algorithms. They showed that leveraging prior expert demonstrations for pre-training can reduce the training time to reach at least the same level of performance compared to plain \gls{rl} by a factor of five. 
Moreover, while they did not compare to classical approaches, they did admit that they do not recommend replacing classical global planning approaches with machine-learning based systems. 
Xie, et al. \cite{xie2018learning} used a PID controller, instead of \gls{il}, as a ``training wheel'' to ``kick-start'' the training of \gls{rl}. They showed that not only does this technique train faster, it is also less sensitive to the structure of the \gls{drl} network and consistently outperforms a standard \gls{ddpg} network. They also showed that their learned policy is comparable with \gls{ros} \texttt{move\_base} \cite{ros_move_base}, a state-of-the-art classical navigation stack. 

Based on the initial success achieved by \gls{il}, such as the work by Pfeiffer, et al. \cite{pfeiffer2017perception} and Zhou, et al. \cite{zhou2019towards}, Lin, et al. \cite{lin2019flying} improved upon \gls{il} using \gls{rl} with full access to state information from a MoCap system. Similar performance to \gls{mpc} approaches could be achieved by combining \gls{il} with \gls{rl}, and possibly with lower thrust, showing potential to outperform classical approaches. 

\vspace{8pt}

\subsubsection{Duplication with Alternate Interfaces and Constraints}
\label{sec::different_approach}
Our second categorization of work that uses machine learning in order to solve the classical navigation problem focuses on techniques that use machine learning in order to leverage---or, in some cases, {\em better} leverage---alternate interfaces and constraints. For example, while vision-based navigation may be possible using classical methods and expensive or brittle computer vision modules, machine learning methods for vision-based navigation can sidestep these issues by not relying on such specific modules. More broadly speaking, we have identified three major ways in which learning methods have allowed navigation systems to leverage alternate interfaces and constraints: (1) using visual input, (2) allowing for complex dynamics, and (3) allowing execution on low-cost hardware.

\vspace{6pt}

\paragraph{Visual Input} 
The ability to effectively use visual input is one of the most important reasons that researchers have sought to use machine learning for motion planning and control in mobile navigation. Although there exist modular approaches for integrating visual information with classical methods (e.g., optical flow, visual odometry, visual \gls{slam}, etc.), these methods usually depend on human-engineered features, incur a high computational overhead, and/or are brittle in the face of environmental variations. Machine learning methods, on the other hand, provide an alternate pathway by which system designers can incorporate visual information.

Among the first attempts to use machine learning for vision-based navigation, \gls{alvinn} \cite{pomerleau1989alvinn}, and DARPA's \gls{lagr} program \cite{muller2006off} proposed end-to-end machine learning approaches to map images directly to motion commands. More recently, NVIDIA's DAVE-2 system \cite{bojarski2016end} represents the culmination of these approaches, leveraging deep learning, large amounts of data, and access to a vast amount of computing power for training. These end-to-end approaches circumvent typical, complicated visual processing pipelines in classical navigation, and have been empirically shown to enable moving-goal navigation capabilities like lane keeping and obstacle avoidance. Codevilla, et al. \cite{codevilla2018end} proposed a similar approach, but the system allows for fixed-goal navigation using conditional \gls{il}. Unfortunately, it is difficult to assess just how much better these approaches are able to leverage visual information for navigation over their classical counterparts because no such experimental comparison has yet been performed.

The diverse and informative visual features in indoor environments has led to a large body of literature that studies using machine learning to incorporate visual information for the particular problem of indoor navigation. This line of research has mostly existed within the computer vision community, and it has focused more on visual perception as opposed to motion planning and control. For example, Gupta, et al. \cite{gupta2017cognitive} proposed to bypass visual \gls{slam} for indoor navigation by incorporating a latent representation within a \gls{vin} and trained the system end-to-end. 
In their later work \cite{gupta2017unifying}, they incorporated noisy but very simple motion controls, i.e., moving forward failed with 20\% probability.
The agent stays in place when the action fails.
Sepulveda, et al. \cite{sepulveda2018deep} used \gls{il} to learn indoor visual navigation with \gls{rrt}* \cite{karaman2011sampling} as expert and utilized indoor semantic structures. 
Zhu, et al. \cite{zhu2017target} used \gls{rl} for visual indoor navigation, where---in addition to bypassing classical visual input processing---they even allowed for the goal to be specified as an image which the robot is expected to see when it reaches the goal. 
One-shot \gls{rl} with interactive replay for visual indoor navigation was proposed by \cite{bruce2017one}, in which the training environment was built using a single traversal with a 360\degree~camera and only the same goal used for training can be reached during deployment. Again, these works did not compare to classical approaches.

Moving from computer vision towards the robotics community, Intention-Net \cite{gao2017intention} kept the classical global and local navigation hierarchical architecture, and served to replace a local planner. It processed visual inputs using a neural network and computed navigation actions without explicit reasoning. An interesting point is that they also represented the local goal as a path image generated from the global planner. 
Alternatively, Kahn, et al. \cite{kahn2018self} proposed a technique based on generalized computation graphs through which a robot could learn to drive in a hallway with vision from scratch using a moving-goal approach. They compared their approach with many learning methods, but not to any classical approaches. 
Kahn, et al. \cite{kahn2018composable} further collected event cues during off-policy exploration and then allowed different navigation tasks without re-training during deployment, with onboard RGB camera. Again, they did not compare to any classical approaches. 
Richter and Roy \cite{richter2017safe} used rich contextual information from visual cameras to predict collision and an autoencoder to decide if the prediction is trustworthy. In case of novel visual perception, they revert back to a safe prior estimate. 

Finally, in the aerial domain, mapping directly from vision to action without any classical visual processing has also been investigated. Ross, et al. \cite{ross2013learning} used \gls{dagger} to map vision directly to \gls{uav} steering commands in order to avoid trees, without any other visual processing. Similar end-to-end visual navigation was performed by Giusti, et al. \cite{giusti2015machine} using supervised data from a hiker as a classification problem to fly on forest trails. Loquercio, et al. \cite{loquercio2018dronet} obtained their flying training set from driving data and trained DroNet. Sadeghi and Levine \cite{sadeghi2016cad2rl} removed the necessity of supervised action labels for vision and used \gls{rl} from simulated visual input. They then directly applied the policy to the real world. Again, however, none of these approaches were evaluated against their more classical counterparts.

\vspace{6pt}

\paragraph{Dynamics Modeling}
Just as machine learning has been used as a promising new pathway by which to exploit visual information for navigation, researchers have also studied using machine learning methods for navigation as an alternative way to handle complex dynamics. Classical dynamics modeling approaches can be very difficult to implement for reasons such as sensing and actuation uncertainty, environmental stochasticity, and partial observability. Learning methods, on the other hand, can bypass these difficulties by directly leveraging data from the platform.

One such approach is that proposed by Zhang, et al. \cite{zhang2016learning}, who addressed the problem of partial observability of \gls{uav} obstacle avoidance by training a system with access to full state information and deployment with only partial observations. Faust, et al. \cite{faust2018prm} used a \gls{rl} agent to bypass the modeling of local \gls{mpc} with non-trivial dynamics, i.e., differential drive ground robot and aerial cargo delivery with load displacement constraints. Becker, et al. \cite{becker2020learning} eliminated the entire modeling of flight dynamics, and used \gls{rl} to ``learn to fly'' without any obstacles. The only comparison to classical approaches was done by Faust, et al. \cite{faust2018prm}, but it was to a trivial straight-line local planner, instead of a better alternative. 

\vspace{6pt}

\paragraph{Low-Cost Hardware}
As a final way in which the autonomous navigation research community has used machine learning approaches to leverage alternate interfaces and constraints, some approaches have sought to use learned systems as a way to relax the burdensome hardware constraints sometimes imposed by more classical systems. For example, Tai, et al. \cite{tai2017virtual} trained a map-less navigator using only 10 laser beams from the LiDAR input for the purpose of enabling navigation using very low-cost sensing hardware (e.g., sonar or ultrasonic arrays instead of high-resolution LiDAR sensors). Additionally, Pan, et al. \cite{pan2020imitation} used supervised training data obtained by a sophisticated \gls{mpc} with high-precision sensors (IMU and GPS), and then trained an off-road vehicle with low-cost sensors (camera and wheel speed sensor) to perform agile maneuvers on a dirt track. They showed that they could achieve similar performance as the \gls{mpc}-expert even though the platform was less well-equipped.

\vspace{8pt}

\subsubsection{Improvement}
\label{sec::improvement}
In this section, we discuss machine learning approaches for autonomous navigation that have explicitly sought to improve over classical approaches.

One set of approaches adopts the approach of automatically tuning the parameters of classical navigation systems. Traditionally, these classical systems are tuned by hand in order to achieve good performance in a particular environment, and the tuning process must be done by a human expert who is intimately familiar with the inner workings of the particular planning components in use. Unfortunately, this process can prove to be tedious and time consuming---the tuning strategy often reduces to one of trial and error. However, finding the correct parameterization can often result in drastically-improved performance. Therefore, one of the main thrusts of machine learning techniques that seek to provide improved navigation performance is toward techniques that can automatically find good parameterizations of classical systems. \gls{il} is a typical learning paradigm used here, as demonstrations---even from non-experts---have been shown to provide a good learning signal for parameter tuning. Bhardwaj, et al. \cite{bhardwaj2019differentiable} used demonstrations and a differentiable extension to a differentiable global planner (\gls{gpmp2}) to find the appropriate obstacle covariance parameter for their planner, while Teso, et al. \cite{teso2019predictive} used \gls{anfis} and backpropagation to find the right set of optimization weights for \gls{dwa} based on demonstration. While we expect that the parameters found through the learning procedure would achieve better navigation performance than random or default parameters, no experiments were actually performed on a real robot.
\gls{appl} by Xiao, et al. \cite{xiao2021appl} took this paradigm a step further and demonstrated the possibility of dynamically changing parameters during deployment to adapt to the current environment. \gls{appl} parameter policies were learned from different human interaction modalities and achieved improved navigation performance compared to classical approaches under one single set of parameters, either default or manually tuned, on two robot platforms running different local planners. Instead of learning adaptive planner parameters to improve navigation performance, learning adaptive motion primitives \cite{sood2020learning} in conjunction with classical planners has effectively reduced computation time.

Another machine learning approach that explicitly seeks to improve navigation performance over classical approaches is PRM-AutoRL by Chiang, et al. \cite{chiang2019learning}. AutoRL automatically searches for an appropriate \gls{rl} reward and neural architecture. While the authors' previous PRM-RL work \cite{faust2018prm} suffered from inferior results compared to a classical system (PRM-DWA), it was claimed that after fine-tuning PRM-RL, PRM-AutoRL was able to outperform the classical approach. However, no similar effort was made to fine tune the classical approach. 
The \gls{lfh} framework by Xiao, et al. \cite{xiao2020toward} also aimed at improving the agility of classical navigation, especially in highly-constrained spaces where sampling-based techniques are brittle. Superior navigational performance is achieved compared to default \gls{dwa} planner, and even to fined-tuned \gls{dwa} by \gls{appl} \cite{xiao2020toward} and to \gls{bc} \cite{pfeiffer2017perception}, both from human demonstration. The Lifelong Navigation by Liu, et al. \cite{liu2020lifelong} claimed it is unnecessary to learn the entire local planner, but only a complement planner that takes control when the classical approach fails. They also addressed the catastrophic forgetting problem for the neural network based planner. Default \gls{dwa} used in conjunction with the learned controller performs better than both classical and learning baselines. 

Lastly, for the particular task of navigating in structured but unknown environments, Stein, et al. \cite{stein2018learning} proposed a machine learning approach for determining path costs on the basis of structural features (e.g., hallways usually lead to goals, rooms are likely to result in a dead end, etc.). While defining costs associated with a particular subgoal (boundary point between known and unknown) is very difficult to do manually, the authors showed that it was very tractable to do so using past experience. The resulting Learned Subgloal Planner \cite{stein2018learning} was shown to be able to outperform a classical optimistic planner. 

\vspace{10pt}

\subsection{Learning beyond Classical Navigation}
\label{sec::beyond_classical}
While most classical approaches are already good at goal-oriented, geometric-based, collision-free navigation, more intelligent mobile robots require more complex navigational behaviors. 
While there have been several attempts made to formulate solutions to these problems using the framework of the classical navigation system over the years, the additional complexity of these problems has recently motivated the research community to consider learning-based solutions as well. The particular complex behaviors that the learning community has focused on are (1) non-geometric terrain-based navigation (Section \ref{sec::terrain_based}), and (2) interactive social navigation with the presence of other agents, either robots or humans (Section \ref{sec::social_navigation}).

\vspace{8pt}

\subsubsection{Terrain-Based Navigation}
\label{sec::terrain_based}
One capability beyond classical geometric-based navigation is terrain-based navigation, i.e., the ability to model navigation behavior on the basis of the physical properties of the ground with which the robot interacts. Unlike geometric information, terrain-related navigation costs are hard to hand-craft in advance, but recent research in machine learning for navigation has shown that it may be easier to learn such behaviors using human demonstration and experience interacting with the environment.

For example, Siva, et al. \cite{siva2019robot} used a human demonstration of navigation on different types of outdoor unstructured terrain, and then combined representation learning and \gls{il} to generate appropriate motion on specific terrain.
Using \gls{irl}, Wigness, et al. \cite{wigness2018robot} learned cost functions to enable outdoor human-like navigational behaviors, such as navigating on a road but staying close to grass, or keeping out of more visible and open areas.
Another approach for terrain-based navigation leveraged real interactions with the physical world to discover non-geometric features for navigation \cite{kahn2020badgr}: traversing over tall grass has very low cost, while the cost of going through uneven terrain is relatively high. 
The learned inverse kindodynamics model conditioned on inertia embeddings by Xiao, et al. \cite{xiao2021learning} was able to capture unknown world states and therefore enable accurate, high-speed, off-road navigation.

\vspace{8pt}

\subsubsection{Social Navigation}
\label{sec::social_navigation}
Going beyond classical single-robot navigation in a relatively static world, several learning methods have sought to enable autonomous social navigation, i.e. navigation in the presence of other agents, either robots or humans. Using classical methods in a dynamic world with other agents is difficult because (1) cost function definitions are usually elusive, e.g., due to different social norms, and (2) such navigation may also include compounding sequential interactions between the agents which are hard to anticipate in advance, i.e. the robot's navigation decision at one time instant will often influence the navigation behavior of another agent at the next instant, which may affect robot's next navigation decision, and so on. As a means by which to address these difficulties, the research community has proposed learning-based solutions.

\vspace{6pt}

\paragraph{Learning Cost Functions}
The difficulties encountered in specifying appropriate static cost functions for social navigation scenarios has motivated the research community to look at using learning-based methods to find these cost functions instead.
For example, in the presence of humans, traditional strategies for defining a costmap (e.g., predefined costs to induce obstacle avoidance) becomes clumsy because humans are not typical obstacles---robots should be allowed to come closer to humans in crowded spaces or when the robot and a human are both moving in the same direction. 

While it may be difficult to manually define a cost function for social scenarios, experiments have shown that it is relatively easy for humans to demonstrate the desired navigation behavior. Therefore, as a first approach, a few research teams have proposed approaches that have adopted \gls{il} as a strategy for learning social navigation behaviors. Tai, et al. \cite{tai2018socially} used \gls{gail} to sidestep the problem of learning an explicit cost function, and instead directly learn a navigation policy that imitates a human performing social navigation. Compared to a baseline policy learned using a simpler \gls{il} approach, their system exhibited safer and more efficient behavior among pedestrians. Yao, et al. \cite{yao2019following} also proposed a similar approach that bypassed cost representation through adversarial imitation learning by using Group-Navi GAN to learn how to assign local waypoints for a local planner. 

Researchers have also investigated methods that seek to learn more explicit cost representations using \gls{irl}. The most straightforward application of \gls{irl} to the problem of social navigation was the approach proposed by Kim and Pineau \cite{kim2016socially} that learns a local cost function that respects social variables over features extracted from a RGB-D sensor. Additionally, Henry, et al. \cite{henry2010learning} used \gls{irl} to learn global cost function representation. Okal and Arras \cite{okal2016learning} developed Trajectory Bayesian \gls{irl} to learn cost functions in social contexts as well. The work by Perez, et al. \cite{perez2018teaching} and Shiarlis, et al. \cite{shiarlis2017rapidly} also taught the robot a cost function with \gls{irl} and then used \gls{rrt} to plan. 

Although not directly using \gls{irl}, Luber, et al. \cite{luber2012socially} learned a costmap from human motion prototypes extracted from publicly available surveillance data. Johnson and Kuipers \cite{johnson2018socially} formed social norm cost in addition to obstacle cost from exploration data collected by a classical planner. The predicted path learned from demonstrations by Perez, et al. \cite{perez2018learning} can also serve as a costmap and partially bias the configuration space sampling of the \gls{rrt} planner. 
Machine learning approaches for social navigation that use learned costs more implicitly include the maximum entropy probability distribution approach from Pfeiffer, et al. \cite{pfeiffer2016predicting}, which was used to model agents’ trajectories for planning; and the approach from Kretzschmar, et al. \cite{kretzschmar2016socially}, which sought to infer the parameters of the navigation model that best matched observed behavior.
Lastly, ClusterNav \cite{martins2019clusternav} clustered human demonstrations to generate a pose graph, which was used to define favorable social behaviors for navigation. All the vertices on this graph were deemed with acceptable cost in the social context. Navigation behaviors were generated using A* to search on this graph to find a path with minimum distance. 

\vspace{6pt}

\paragraph{Learning Controllers}
Another challenge that arises when trying to use classical navigation approaches for social navigation is the inherent difficulty in modeling the compounding effects of sequential interactions between agents. Most classical approaches are designed only to operate within a static world, and, even when they are contrived to handle dynamic situations, it becomes extremely difficult to encode additional rules about how current actions will affect other agents, and how the decisions made by those agents might impact future decisions. As an alternative approach, \gls{rl} algorithms have been shown to provide a suitable machine learning framework that can capture the essence of such sequential social interactions.

For example, Chen, et al. \cite{chen2017decentralized} proposed CADRL, a deep-learning approach for learning controllers capable of navigating multiple robots to their goals in a collision-free manner. Later, they proposed a variant of the CADRL framework \cite{chen2017socially} which uses \gls{rl} with a hand-crafted reward function to find navigation systems that incorporate social norms such as left- and right-handed passing. Everett, et al. \cite{everett2018motion} augmented this framework with an \gls{lstm} \cite{hochreiter1997long} and GPU (GA3C-CADRL), both of which allow an \gls{rl} algorithm to successfully learn navigation policies that can handle the complex sequential interaction between the robot and other pedestrians. Similarly, the ego-safety and social-safety components of the reward function designed by Jin, et al. \cite{jin2019mapless} also allowed an \gls{rl} approach to capture sequential interaction. The \gls{scn} problem considered by Lin, et al. \cite{li2018role} additionally required the robot to maintain close to a companion while travelling together towards a certain goal in a social context, which increased the complexity of the sequential interactions between agents. Role-playing learning \cite{li2018role} using \gls{rl} also aimed at resolving this additional difficulty. 

Utilizing \gls{rl}-based approaches to capture compounding sequential interactions, researchers have shown that learning methods can achieve better performance than multi-robot navigation approaches that do not (or cannot) consider such interactions, e.g., \gls{orca} \cite{van2011reciprocal}, which is a state-of-the-art multi-robot collision avoidance strategy that doesn't pay explicit attention to social awareness. 
For example, in simulation, Godoy, et al. \cite{godoy2018alan} used \gls{rl} to generate preferred velocity and then passed to \gls{orca} to produce collision-free velocity. 
Also in simulation,  Chen, et al. \cite{chen2017decentralized} (CADRL) exhibited a 26\% improvement in time to reach the goal can be achieved compared to \gls{orca}. 
Long, et al. \cite{long2018towards} also used \gls{drl} with hand-crafted reward and curriculum learning for decentralized multi-robot collision avoidance, and---compared to non-holonomic \gls{orca}---the approach achieved significant improvement in terms of success rate, average extra time, and travel speed, with up to 100 robots. 
Liang, et al. \cite{liang2020crowdsteer} further showed their collision avoidance policy using multi-sensor fusion and \gls{drl} can achieve better performance than \gls{ros} \texttt{move\_base} \cite{ros_move_base} and the work by Long, et al. \cite{long2018towards} in social navigation. 
Before planning motion, Chen, et al. \cite{chen2019crowd} used \gls{rl} to model both human-robot and human-human interactions and their relative importance, and they showed that the resulting system outperformed \gls{orca}, CADRL, and GA3C-CADRL.
Finally, a hierarchical \gls{rl} framework with a \gls{hmm} model to arbitrate between target pursuit and a learned collision avoidance policy \cite{ding2018hierarchical} also achieved better results than \gls{orca} and the work by Long, et al. \cite{long2018towards}, indicating that unwrapping the end-to-end learning black box based on human heuristics may improve performance.
Different from an end-to-end perspective, the local trajectory planner and velocity controller learned in Pokle, et al. \cite{pokle2019deep} uses attention to balance goal pursuit, obstacle avoidance, and social context. It can achieve more consistent performance compared to \gls{ros} \texttt{move\_base} \cite{ros_move_base}.

\vspace{10pt}

\subsection{Analysis}
Having organized the machine learning for autonomous navigation literature according to similarities and differences with classical autonomous navigation approaches, we now provide a summary analysis. Most importantly, we find that, while a large concentration of learning-based work is able to solve the classical navigation problem, very few approaches actually {\em improve} upon classical techniques: currently, for navigation as a robotics task, learning methods have been used mainly to replicate what is already achievable by classical approaches; and for learning, navigation is mostly a good testbed to showcase and to improve learning algorithms. 
Additionally, a number of learning-based techniques have been proposed that have enabled relatively new navigation capabilities such as terrain-based and social navigation, which have proven difficult to achieve with classical techniques.

Specifically, as illustrated in Figure \ref{tab::comparison}:
\begin{enumerate}
    \item \textbf{While the majority of papers surveyed sought to solve the classical navigation problem, very few actually demonstrate improved performance over classical approaches.} 46 of the 74 papers surveyed dealt with the classical problem, and only eight of those showed improved performance over classical solutions. The other 38 only achieved navigation in relatively simple environments (e.g., sparse obstacles, uniform topology, same training and deployment environment) and mostly did not compare with classical approaches. One explanation is that the research community initially focused on answering the question of whether or not navigation was even possible with learning approaches, and focused on the unique benefits of such approaches in that extensive human engineering (e.g., filtering, designing, modeling, tuning, etc.) is not required or that alternate interfaces and constraints can be leveraged. Very few proposed learning approaches have been compared to classical solutions, which leads us to advocate for such performance comparison to be done in future work, along with comparisons along the other dimensions, such as training vs. engineering effort and neural architecture search vs. parameter tuning (items 2 in Section \ref{sec::directions}). 
    
    \item \textbf{Of the papers surveyed that sought to enable navigation behavior beyond that of the classical formulation, the majority focus on social navigation and only a few on terrain-based navigation.} In particular, 28 out of the 74 papers surveyed sought to go beyond classical navigation, and 24 of those focused on social navigation. The other four sought to enable terrain-based navigation. Social navigation may be the predominant focus due to the relative ease with which the problem can be studied, i.e., humans and pedestrian environments are much more easily accessible to researchers than environments with challenging terrain. 
    
\end{enumerate}

\section{OTHER TAXONOMIES}
\label{sec::taxanomies}
In this section, we review the literature covered above in the context of two additional taxonomies: 
(1) the specific navigation task that is learned, and (2) the input modalities to the learning methods. Rather than describing each paper as in the previous sections, we will only define the categories below, and visually indicate how the papers fall into the categories via Figures \ref{tab::tasks} and \ref{tab::modalities}. Note that Sections \ref{sec::learning} and \ref{sec::comparison} discussed some of the literature according to these taxonomies, but only as a way to group a subset of reviewed papers in their primary categories. Here, however, we categorize each of the reviewed papers using these two taxonomies. Note that Figures \ref{tab::tasks} and \ref{tab::modalities} only include the 74 papers selected within the specific scope described in Section \ref{sec::introduction} (i.e.,  machine learning approaches for motion planning and control that actually move mobile robots in their environments), and are not exhaustive in terms of using machine learning in their particular categories in general.

\vspace{10pt}

\subsection{Navigational Tasks}
\label{sec::navigational_tasks}
\begin{figure*}
\centering
\caption{Navigational Tasks}
\label{tab::tasks}
\includegraphics[width=\textwidth]{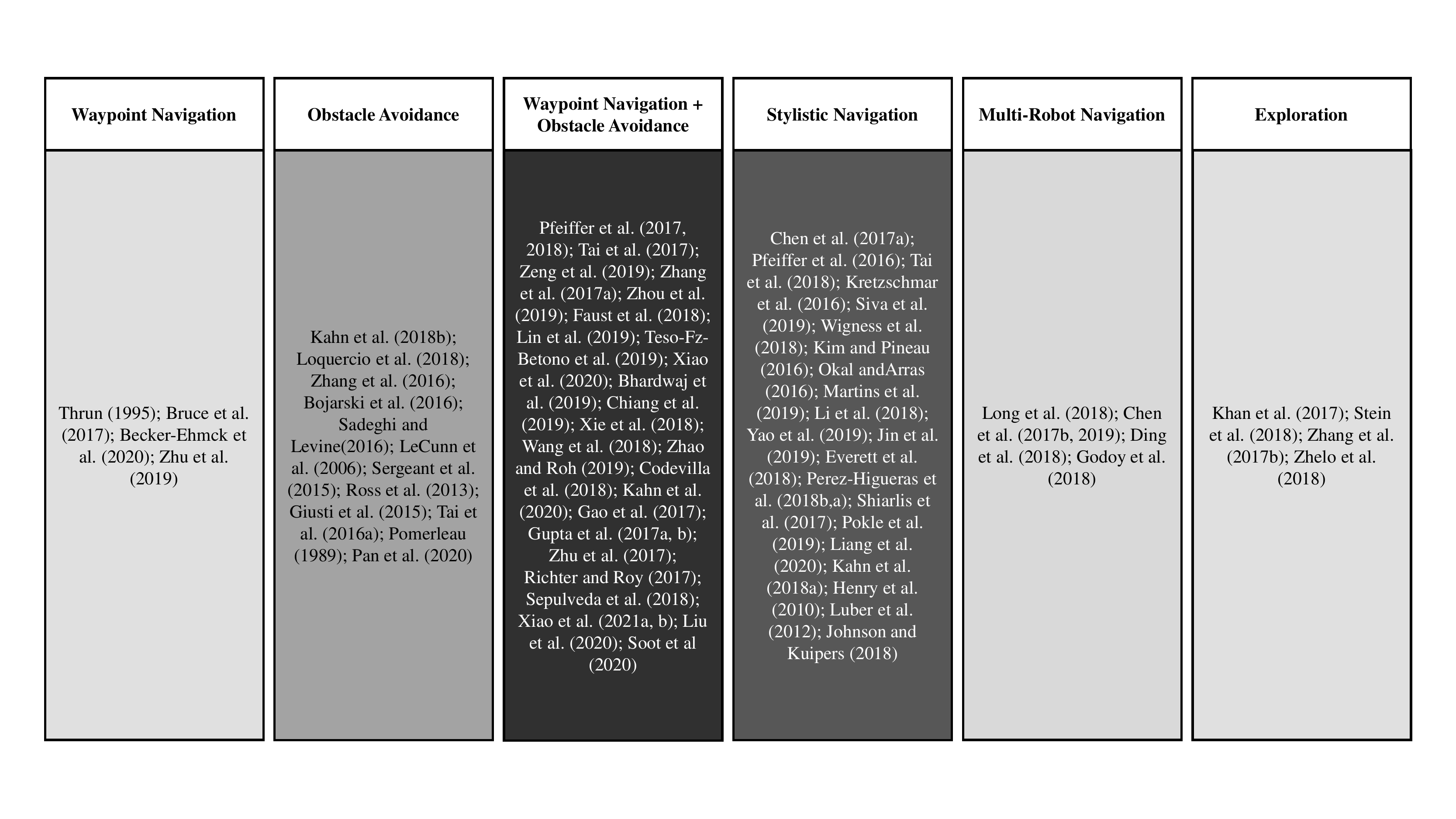}
\end{figure*}

We have identified six different navigational tasks considered by the literature on machine learning for motion planning and control in mobile robot navigation. They are (1) \emph{Waypoint Navigation}, (2) \emph{Obstacle Avoidance}, (3) \emph{Waypoint Navigation + Obstacle Avoidance}, (4) \emph{Stylistic Navigation}, (5) \emph{Multi-Robot Navigation}, and (6) \emph{Exploration}. All the reviewed papers and their task categories are summarized in Figure \ref{tab::tasks}. 

\vspace{8pt}

\subsubsection{Waypoint Navigation}
The task of waypoint navigation is a fundamental building block for mobile navigation. The task definition is to move the robot to a specified goal, without considering any other constraints, e.g., obstacle avoidance. Works in this category are few, and are mostly useful as a ``proof-of-concept'' for learning methods since real-world requirements for navigation typically require taking into account additional constraints.

\vspace{8pt}

\subsubsection{Obstacle Avoidance}
Another important building block of navigation is the task of obstacle avoidance, which is critical to the issue of safety in mobile robotics. The works we've placed into this category consider {\em only} obstacle avoidance, and do not navigate to a pre-specified goal. Note that all Moving-Goal Navigation approaches in Section \ref{sec::moving_goal} fall into this category, and here we apply this categorization across {\em all} the papers we have surveyed. The specific task of hallway navigation is a representative example of obstacle avoidance. 

\vspace{8pt}

\subsubsection{Waypoint Navigation + Obstacle Avoidance}
Waypoint navigation with obstacle avoidance is the task in our taxonomy that most closely matches that of classical navigation: the task requires the robot to navigate to a specified goal while avoiding collisions with obstacles along the way. The majority of learning-based motion planning and control approaches for mobile robot navigation belong here. Note the majority of works in Fixed-Goal Navigation in Section \ref{sec::fixed_goal} belong to the Waypoint Navigation + Obstacle Avoidance category in this taxonomy. Again, here we apply this categorization across {\em all} the papers.

\vspace{8pt}

\subsubsection{Stylistic Navigation}
Stylistic navigation is a more complex and varied task compared to those above; it refers to generating any number of navigation behaviors that systematically differ from minimum-cost, collision-free navigation, and that are often defined with respect to features in the environment beyond geometric obstacles. Examples of stylistic navigation include staying to one side of the hallway as in social navigation, and moving slower or faster depending on the type of terrain being traversed. Stylistic navigation tasks are not typically easily addressed by classical approaches, and because of the elusive nature of explicitly defining such tasks using classical static cost representations, they have mostly been enabled using \gls{il} and \gls{rl} methods.

\vspace{8pt}

\subsubsection{Multi-Robot Navigation}
The multi-robot navigation task requires an agent to explicitly take into account the presence of other navigating agents when deciding its own course of action. Classical approaches to multi-robot navigation typically require hand-crafted, heuristic behaviors, which quickly become insufficient as more and more agents need to be considered. Machine learning approaches to this task, on the other hand, are able to utilize experiential data to find successful policies. 

\vspace{8pt}

\subsubsection{Exploration}
The last task we consider for autonomously navigating mobile robots is that of pure exploration, i.e., when the navigating agent's goal is to maximize coverage of an environment for the purposes of, e.g., mapping or surveillance. For this task, machine learning techniques have been used to perform mapping (by, e.g., memorizing an explored space), or to help form predictions over unknown space based on previous training experience.

\vspace{10pt}

\subsection{Input Modalities}
\label{sec::input_modalities}
\begin{figure*}
\centering
\caption{Input Modalities}
\label{tab::modalities}
\centering
\includegraphics[width=\textwidth]{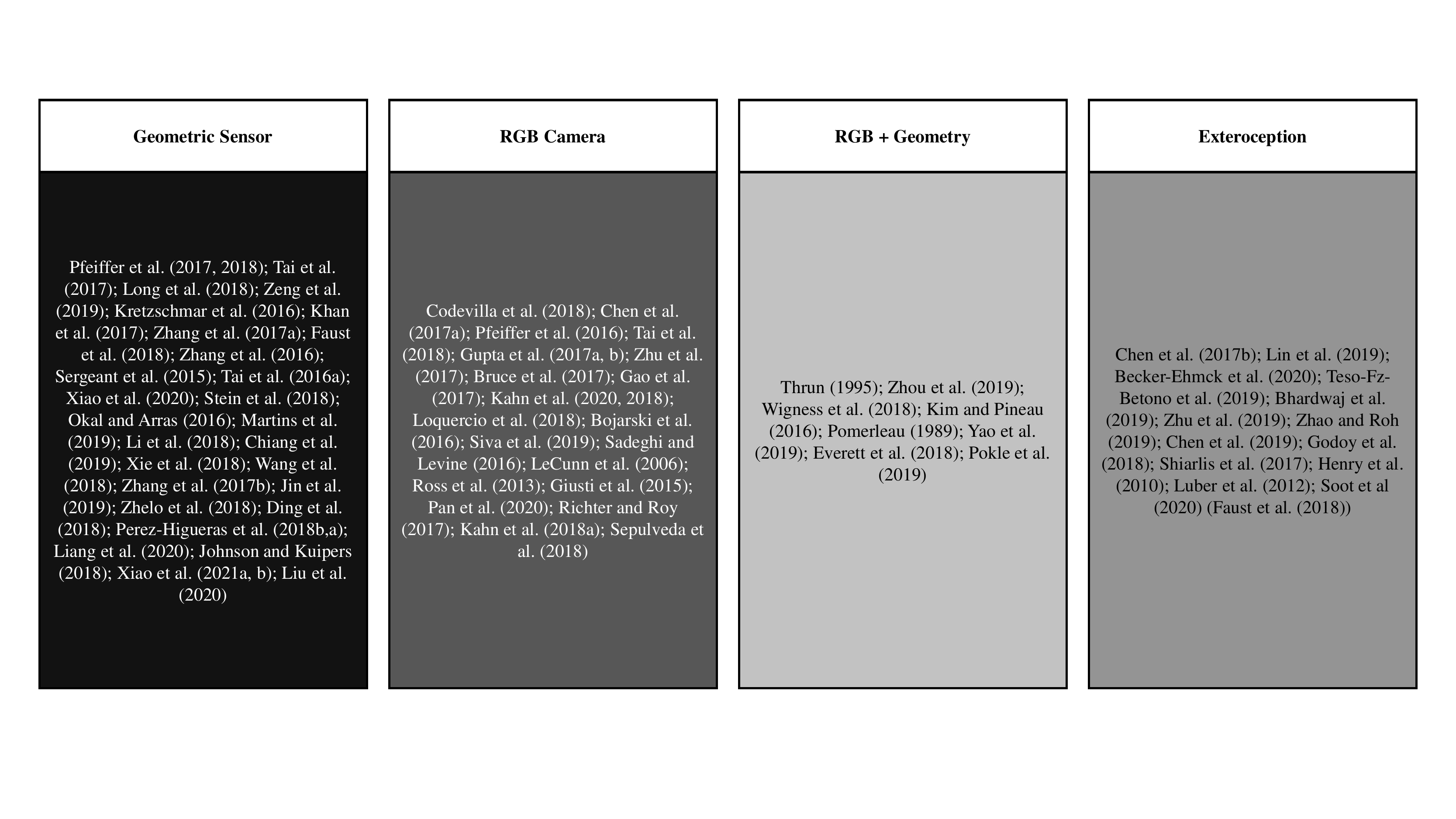}
\end{figure*}

Finally, we also organize the literature reviewed in this survey into categories that consider the system's sensor input modality. In this regard, each of the reviewed learning approaches uses one of the following four classes of input: (1) \emph{Geometric Sensors}, (2) \emph{RGB Cameras}, (3) \emph{RGB + Geometry}, and (4) \emph{Exteroception}. The papers grouped by their input modalities can be found in Figure. \ref{tab::modalities}.

\vspace{8pt}

\subsubsection{Geometric Sensors}
Classically, the autonomous navigation problem has been posed in a purely geometric sense, i.e., the task is to move through a geometric workspace to a goal located at a set of specified coordinates while avoiding obstacles of different sizes and shapes along the way. In this sense, the only geometric information needed is the location of generic obstacles within the environment. Therefore, sensors that can directly provide this information are widely used to provide perception for many navigation systems, e.g., LiDAR sensors, depth sensors, sonars, etc. This input modality categorization contains the largest number of learning-based approaches in our survey.

\vspace{8pt}

\subsubsection{RGB Cameras}
RGB images lack depth information, and therefore most classical navigation approaches required at least stereo vision for depth triangulation so as to derive geometric information. Learning methods have relaxed this requirement; most learning methods require only monocular vision to compute motion for navigation. Although precise geometric information is not possible with only a single camera, reasonable motion, e.g., lane keeping, turning left when facing a tree, etc., can be effectively generated. RGB cameras are low-cost, are human-friendly, and can provide rich semantic information, which has drawn the attention of the learning community.

\vspace{8pt}

\subsubsection{RGB + Geometry}
The third categorization in this section includes works that combine the advantages of both RGB and geometric perception. Geometric sensors are typically used to enable obstacle avoidance, while visual cameras provide semantic information.

\vspace{8pt}

\subsubsection{Exteroception}
Work in this final category, i.e., methods that utilize input from external sensors, are typically only used in controlled environments such as a \gls{mocap} studio or a simulator. These methods assume perfect sensing and perception, and focus instead on what happens next, such as learning a model or policy, finding the right parameterization, etc. 

\vspace{10pt}

The taxonomies proposed in this section are intended to give readers an overview of what are the navigational tasks researchers have used machine learning for, and the sensor modalities proposed learning approaches utilize as input. For the navigational task, although goal-oriented navigation with obstacle avoidance still comprises the largest focus, learning methods have drawn attention to interactive navigation in terms of navigation style and multi-robot navigation. Learning methods have also given rise to many navigation systems that can utilize visual---as opposed to purely geometric---input sensors, which represents a significant departure from the classical literature.

\vspace{12pt}

\section{ANALYSIS AND FUTURE DIRECTIONS}
\label{sec::insights}
We now present an analysis of the literature surveyed above.
We will discuss our findings based on not only each individual organization proposed in the previous sections, but also from a perspective that unifies the proposed taxonomies.
Then we provide suggestions for future research directions.

\vspace{10pt}

\subsection{Analysis}
\label{sec::analysis}
We first recap the statistics in Sections \ref{sec::learning} and \ref{sec::comparison}, and then provide a cross-dimensional analysis between these two sections.

\subsubsection{Recap}

Figure \ref{tab::learning_scope} provides an overview of the scope of machine learning approaches for navigation with respect to the classical navigation pipeline.
We note that 43 out of the 74 surveyed papers are end-to-end approaches, and 15 of those lack the ability to navigate to user-defined goals.
This inability to navigate to defined goals is not common in the classical navigation literature. 
The remaining 31 papers describe approaches that are {\em not} end-to-end, and each of these {\em can} navigate to user-defined goals: 13 applied machine learning to a particular navigation subsystem---one for global planning and 12 for local planning---and the other 18 applied machine learning to individual components of a classical navigation system---14 for representation and four for parameters.

Analyzing the literature from a different perspective, Figure \ref{tab::comparison} provides a comparison between learning-based and classical approaches to the autonomous navigation problem with respect to the particular problems and goals that are considered.
First, we find that 46 out of the 74 surveyed learning-based approaches consider the classical navigation problem, but a majority of those (38/46) resulted in navigation systems that were only tested in relatively simple environments and that did not outperform---or, often, were not even compared with---classical approaches.
That said, a small number of these approaches (8/46) {\em were} compared to classical approaches and demonstrated some improvement over them.
Second, the remaining 28 out of the 74 surveyed learning-based papers have proposed techniques for which the goal is to provide capabilities that go {\em beyond} that envisioned by classical autonomous navigation.
Of these 28 papers, 24 sought to enable some form of social navigation, and the remaining four focused on building systems that could perform terrain-based navigation.

\subsubsection{Cross-Dimensional Analysis}
\label{sec::cross_dim}
\begin{figure*}
\centering
\caption{Performance vs. Scope} 
\label{tab::performance_vs_scope}
\rotatebox{90}{\includegraphics[width=1\textheight]{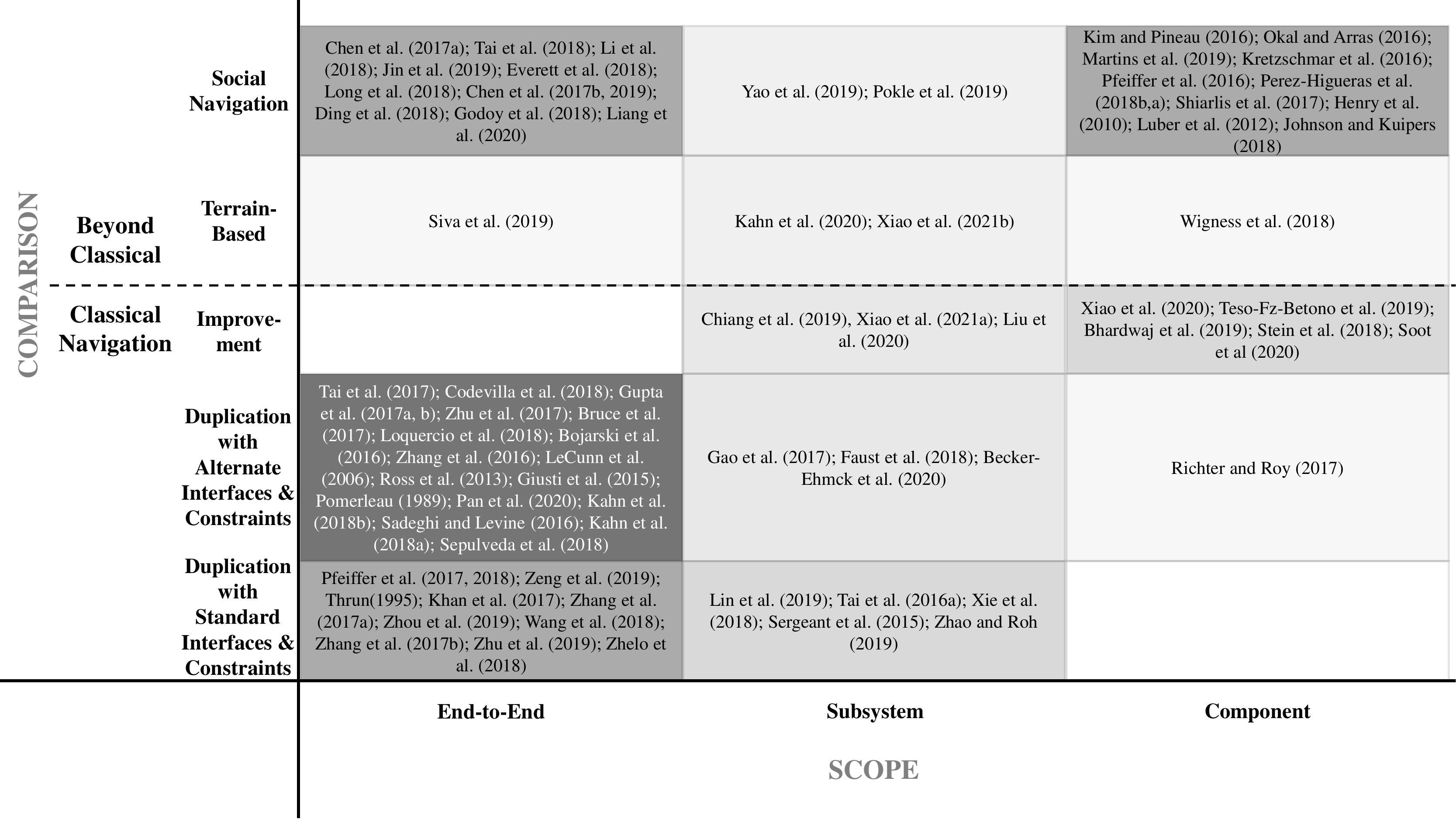}}
\end{figure*}

Using the categorizations discussed in Sections \ref{sec::learning} and \ref{sec::comparison} (i.e., scope of learning and comparison to classical approaches, respectively), we now analyze the relationship between them via the cross-dimensional view we provide in Figure \ref{tab::performance_vs_scope}.

First, consider the upper portion of Figure \ref{tab::performance_vs_scope}, which focuses on the machine learning literature that has sought to extend the capabilities of navigation systems beyond what was classically envisioned.
In particular, one can see that, while many learning-based approaches have been proposed for social navigation, the scope of these approaches has either been very broad (we classify 11 works as end-to-end) or very narrow (an additional 11 works are classified as component-level, and each of them focuses specifically on learning cost function representations).
In fact, only two approaches that consider social navigation have been proposed at the intermediate subsystem level: \cite{yao2019following} for global planning and \cite{pokle2019deep} for local planning.
The upper portion of the figure also shows that relatively few learning-based approaches have been proposed to enable terrain-based navigation, though they are distributed across our end-to-end \cite{siva2019robot}, subsystem \cite{kahn2020badgr} \cite{xiao2021learning}, and component \cite{wigness2018robot} categories.

Analyzing the lower portion of Figure \ref{tab::performance_vs_scope} also yields some interesting insights.
In particular, one can observe a correlation between the two axes, i.e., approaches that look to apply machine learning to a more limited scope of the classical navigation problem seem to, roughly speaking, yield better results.
On this point, several more-specific observations can be made.
For one, it is interesting to note that all end-to-end approaches fall in either the Duplication with Standard or Alternative Interfaces and Constraints categories, with no single end-to-end approach having actually demonstrated improvement over classical navigation systems.
Subsystem-level approaches, on the other hand, populate each of the categories defined relative to classical navigation systems, with three works in particular \cite{chiang2019learning} \cite{liu2020lifelong} \cite{xiao2020toward} managing to demonstrate improved performance.
In line with the observed correlation stated earlier, learning-based approaches that focus on individual components of a navigation system predominately appear in our Improvement category (with the novelty detection for collision prediction work by Richter and Roy \cite{richter2017safe} as the only exception).
We posit that this correlation exists because, while learning approaches with a wider scope provide an initial qualitative ``proof-of-concept'' that they {\em can} be used for navigation, it may prove too difficult to additionally be able to tune such a large system to achieve better overall performance.
In contrast, learning methods that target a more specific subsystem or component of the navigation pipeline can usually improve the performance of that system because the problem itself is smaller, with the rest of the system relying on reliable components from classical approaches.
The concentration of learning works in the lower left corner of the table suggests that, for classical navigation, the literature has primarily approached the problem in an end-to-end manner and accomplished what has already been achieved by classical methods.

\vspace{10pt}

\subsection{Recommendations and Future Directions}
\label{sec::directions}
Based on the analysis provided in the previous Section \ref{sec::analysis}, we now provide high-level recommendations and identify promising future research directions for using machine learning to improve mobile robot navigation.

\begin{enumerate}
\item \textbf{\textsc{Recommendation}: The current best practice is to use machine learning at the subsystem or component level.}
We make this recommendation based primarily on the correlation we found using the cross-dimensional analysis from Section \ref{sec::cross_dim} (Figure \ref{tab::performance_vs_scope})---the more machine learning approaches limit their focus to particular subsystem or component of a classical navigation system, the better the overall system seems to perform.
Because these best approaches still rely heavily upon several components from classical navigation systems, this observation also serves to remind us of the advantages of those systems.
For example, hybrid systems can employ learning modules to adapt to new environments, and classical components to ensure that the system obeys hard safety constraints.
Below, we highlight two critical aspects of classical systems---{\em safety} and {\em explainability}---that current learning-based approaches do not address, thus motivating their continued inclusion via a hybrid system.

One of the most important system properties that classical navigation system components can provide is a guarantee of {\em safety}, which is typically not present for methods that utilize machine learning only.
That is, unlike classical \gls{mpc} approaches that utilize bounded uncertainty models, learning methods typically cannot provide any explicit assurance of safety.
For mobile robots moving in the real world---where colliding with obstacles, other robots, and even humans would be catastrophic---the importance of safety cannot be over-emphasized.
That machine learning approaches lack safety guarantees is likely one of the most important reasons why they are rarely found in mission-critical systems.
Currently, only 12 out of the 74 surveyed papers describe navigation systems that can provide some form of safety assurance, and each of these does so using a component from a classical system.

Another aspect of classical navigation systems that has been largely overlooked by machine learning approaches is that of {\em explainability}.
Classical methods typically allow human roboticists to log and trace errors back along the navigation pipeline in an understandable and explainable fashion, which allows them to find and resolve specific issues so as to improve future system performance.
However, current learning-based approaches---especially those that rely on deep learning---do not exhibit such explainability.
In fact, none of the 43 end-to-end approaches surveyed even attempted to maintain any notion of explainability, while the other 31 approaches surveyed only partially maintained explainability by leveraging classical navigation subsystems or components.

One concrete example of the type of hybrid system we advocate for as a best practice is one that uses machine learning to tune the parameters of a classical system.
Although classical approaches have the potential to conquer a variety of workspaces enabled by their sophisticated design, when facing a new environment, a great deal of parameter tuning is required to allow it to adapt.
This tuning process is not always intuitive, and therefore requires expert knowledge of the classical navigation system along with trial and error.
Without parameter tuning, a classical navigation system with default parameters generally exhibits only mediocre behavior across a range of environments. 
Only six \cite{teso2019predictive} \cite{bhardwaj2019differentiable} \cite{xiao2021appl} \cite{liu2020lifelong} \cite{sood2020learning} \cite{xiao2021learning} out of the 74 surveyed papers investigated how to improve the adaptivity of classical approaches using machine learning; there is still plenty of work that can be done in this space.

Finally, while we claim that hybrid systems are the current best practice, there exists a large community of researchers interested in navigation systems constructed using end-to-end machine learning.
With respect to these approaches, the literature has shown repeatedly that they can be successful at {\em duplicating} what has already been achieved by classical approaches, but it has not convincingly shown that end-to-end learning approaches can advance the state-of-the-art performance for the navigation task itself.
While some duplication work is indeed necessary in order to provide proofs-of-concept for such systems, we suggest that this community should now move on from that goal, and instead focus explicitly on trying to {\em improve} navigation performance over that afforded by existing systems.
Further, with respect to the aspects of safety and explainability highlighted above as advantages of classical systems, it remains an interesting open problem as to whether learning approaches that exhibit these same features can be designed.

\item \textbf{\textsc{Recommendation}: More complete comparisons of learning-based and classical approaches are needed.}
Across the literature we surveyed, we found that the evaluation methodologies applied to proposed learning-based techniques to be both inconsistent and incomplete.
First, we found it surprisingly common for the experimental evaluations performed for learning-based approaches to omit classical approaches as a baseline.
Second, we found that many papers failed to compare machine learning approaches to classical approaches with respect to several relevant metrics, leading to confusion regarding the relative advantages and disadvantages of each.

With respect to using classical approaches as a baseline, we found that much of the literature that proposed new learning-based approaches primarily chose to limit experimental comparison to other learning-based approaches \cite{zeng2019navigation} \cite{zhou2019towards} \cite{zhelo2018curiosity} and some even only used metrics specific to machine learning concerns, e.g., reward~\cite{zhang2017deep} and  sample efficiency~\cite{pfeiffer2018reinforced}.
However, the fact that one learning algorithm achieves superior performance over another learning algorithm, especially only with respect to learning metrics, is, both practically and scientifically, currently of limited value to the community concerned with autonomous navigation.
Only when learning methods start to outperform the state-of-the-art classical approaches does it make sense to focus on these types of improvements.
Instead, when applying learning methods to navigation problems, we recommend that researchers focus primarily on metrics that measure real-world navigation performance and report where their proposed approaches stand with respect to state-of-the-art techniques from the classical literature.

With respect to comparing to classical systems using additional metrics other than real-world navigation performance, we found that the literature on learning-based navigation insufficiently acknowledges important shortcomings of these approaches.
While the extensive engineering effort required by classical approaches to navigation is often used to motivate the application of machine learning approaches as an alternative (e.g., with classical approaches even attempts to achieve slight performance increases may require substantial knowledge, laborious effort, or expensive hardware and software development), similar costs exist for learning-based approaches. 
However, we found that the literature on learning-based navigation did little to explicitly acknowledge or characterize the very real costs of hyperparameter search and training overhead.
Manual hyperparameter search---e.g., handcrafting reward functions, finding optimal learning rates, searching for appropriate neural architectures---was required for 73 out of 74 of the surveyed approaches to be successful, and the sole remaining work \cite{chiang2019learning} took 12 days to automatically find such parameters.
The costs associated with such searching, which are similar to those incurred by manual rule definition and parameter tuning procedures found when trying to deploy classical methods, should be clearly identified in future work and compared with classical tuning when possible.
Additionally, the data collection and training costs associated with learning methods are rarely made apparent: training data for navigation is difficult to collect, and training is typically done offline using high-end computational hardware over days or even weeks.
For all their shortcomings, classical approaches do not incur such costs.
Therefore, future work in this area should objectively consider this trade-off between engineering costs associated with classical approaches and the costly training overhead associated with learning-based approaches.

\item \textbf{\textsc{Future Direction}: Further development of machine learning methods that target the reactive local planning level of autonomous navigation.}
Based on the work we surveyed, navigation components that implement predominately reactive behaviors seem to have thus far benefited the most from the application of machine learning \cite{xiao2020toward} \cite{liu2020lifelong}  \cite{chiang2019learning} \cite{faust2018prm} \cite{richter2017safe} \cite{xiao2021learning}. 
Intuitively, this success of learning for reactive behaviors is because machine learning approaches are typically at their best when applied to limited, well-defined tasks. 
Local planning, in which we seek to analyze local sensor data to produce immediate outputs over limited spatial windows, is exactly such a task.
Global planning, on the other hand, is typically a much more deliberative process.
Take, as an analogy, human navigation: humans typically perform high-level deliberation to come up with long-range plans, such as how to get from a house or apartment to a local park, but navigation becomes much more reactive at the local level when needing to respond to immediate situations, such as avoiding a running dog or moving through extremely complex or tight spaces.
These reactive behaviors are difficult to model using rule-based symbolic reasoning, but are ripe for learning from experience.
These reasons explain the initial thrust and successes towards learning local planners, especially to address challenging reactive situations.
Only four \cite{richter2017safe} \cite{faust2018prm} \cite{xiao2020toward} \cite{xiao2021learning} out of the 74 surveyed papers have taken the initial step to investigate local navigation with challenging constraints; we believe there is much more work that could be done in this direction. 

\item \textbf{\textsc{Future Direction}: Development of machine learning methods that can enable additional navigation behaviors that are orthogonal to those provided by classical approaches.}
While classical approaches are well-suited to the problem of metric (i.e., minimum-energy, collision-free) navigation in a static environment, our survey has shown that learning-based methods have started to enable qualitatively different types of navigation.
In particular, we found that the community has focused thus far primarily on social navigation, and a bit on terrain-aware navigation.
With these successes, we encourage researchers to investigate other types of navigation behaviors that might now be possible with machine learning.
For example, \gls{appl} \cite{xiao2021appl} has demonstrated the efficacy of dynamic parameter adjustment on the fly, which constitutes a brand new capability for a classical motion planner.
Additionally, researchers may also consider things such as stylistic navigation, navigation behaviors that change in the presence of particular objects, etc.

\item \textbf{\textsc{Future Direction}: Further development of machine learning components that continually improve based on real deployment experience.}
While most traditional navigation systems require manual intervention by human engineers in order to overcome failures or unsatisfactory performance during deployment, 
learning-based systems, in theory, should be able to automatically process data generated by such events and improve without human involvement.
However, most current learning-based systems still separate the training and deployment phases, even for online RL approaches (due to, e.g., onboard computational constraints).
This separation means that such systems are not continually learning from deployment experience.
Only one \cite{liu2020lifelong} of the 74 papers explicitly sought to add this capability and proposed a learning-based system that could improve by continually leveraging actual deployment experience outside of a fixed training phase and entirely onboard a robot. 
This ability to automatically improve during actual deployment using previous successes and failures is one that is not exhibited by classical static approaches to navigation. There is a great opportunity now to design such ``phased'' learning approaches blended in actual deployment for continually improving navigation performance. 

\end{enumerate}

\vspace{12pt}
\section{CONCLUSIONS}
\label{sec::conclusions}

This article has reviewed, in the context of the classical mobile robot navigation pipeline, the literature on machine learning approaches that have been developed for motion planning and control in mobile robot navigation.
The surveyed papers have been organized in four different ways so as to highlight their relationships to the classical navigation literature: the scope of the learning methods within the structured navigation systems (Section \ref{sec::learning}), comparison of the learning methods to what is already achievable using classical approaches (Section \ref{sec::comparison}), navigational tasks (Section \ref{sec::navigational_tasks}), and input modalities (Section \ref{sec::input_modalities}). 
We have discussed each surveyed approach from these different perspectives, and we have presented high-level analyses of the literature as a whole with respect to each perspective.
Additionally, we have provided recommendations for the community and identified promising future research directions in the space of applying machine learning to problems in autonomous navigation (Section \ref{sec::insights}). Overall, while there has been a lot of separate research on classical mobile robot navigation, and, more recently, learning-based approaches to navigation, we find that there remain exciting opportunities for advancing the state-of-the-art by combining these two paradigms.  We expect that by doing so, we will eventually see robots that are able to navigate quickly, smoothly, safely, and reliably through much more constrained, diverse, and challenging environments than is currently possible.

\section{ACKNOWLEDGEMENTS}
This work has taken place in the Learning Agents Research
Group (LARG) at the Artificial Intelligence Laboratory, The University
of Texas at Austin.  LARG research is supported in part by grants from
the National Science Foundation (CPS-1739964, IIS-1724157,
NRI-1925082), the Office of Naval Research (N00014-18-2243), Future of
Life Institute (RFP2-000), Army Research Office (W911NF-19-2-0333),
DARPA, Lockheed Martin, General Motors, and Bosch.  The views and
conclusions contained in this document are those of the authors alone.
Peter Stone serves as the Executive Director of Sony AI America and
receives financial compensation for this work.  The terms of this
arrangement have been reviewed and approved by the University of Texas
at Austin in accordance with its policy on objectivity in research.

We would also like to thank Yifeng Zhu for helpful discussions and suggestions, and Siddharth Rajesh Desai for helping editing and refining the language for this survey. 

\printglossaries

\bibliographystyle{spmpsci}
\bibliography{references.bib}

\end{document}